\documentclass{article}

\usepackage{lmodern}
\usepackage[T1]{fontenc}
\usepackage[left=1.5in,right=1.5in,bottom=1.25in]{geometry}

\usepackage{amsthm}
\usepackage{amsmath}
\usepackage{amsfonts}
\usepackage{authblk}
\usepackage{hyperref}
\usepackage[
  backend=biber,
  natbib=true,
  style=apa,
  sorting=nyt,
  maxcitenames=1,
  url=false, 
  doi=false,
  giveninits=true
]{biblatex}
\usepackage{graphicx}
\usepackage{centernot}
\usepackage{cleveref}
\usepackage{algorithm2e}
\usepackage{todo}
\usepackage{tikz}
\usepackage{wrapfig}
\usepackage{subcaption}
\usepackage{booktabs}
\usepackage{commath}
\usepackage{subcaption}
\usetikzlibrary{bayesnet}
\usetikzlibrary{arrows.meta, positioning, shapes, fit, backgrounds, calc}

\hypersetup{
  colorlinks,
  linkcolor={red!50!black},
  citecolor={blue!50!black},
  urlcolor={blue!50!black}
}
\DeclareNameAlias{author}{last-first}

\title{Simulated-based Inference with the Python Package \texttt{sbijax}}

\author[$\clubsuit$]{Simon Dirmeier\footnote{Correspondence to \texttt{\url{simon.dirmeier@sdsc.ethz.ch}}}}
\author[$\heartsuit$]{Antonietta Mira}
\author[$\diamondsuit$]{Carlo Albert}
\affil[$\clubsuit$]{Swiss Data Science Center and ETH Zurich, Switzerland}
\affil[$\heartsuit$]{Università della Svizzera italiana, Switzerland, and Insubria University, Italy}
\affil[$\diamondsuit$]{Swiss Federal Institute of Aquatic Science and Technology, Switzerland}
\date{}

\begin{document}
\maketitle

\begin{abstract}
Neural simulation-based inference (SBI) describes an emerging family of methods for Bayesian inference with intractable likelihood functions that use neural networks as surrogate models. Here we introduce \texttt{sbijax}, a Python package that implements a wide variety of state-of-the-art methods in neural simulation-based inference using a user-friendly programming interface. \texttt{Sbijax} offers high-level functionality to quickly construct SBI estimators, and compute and visualize posterior distributions with only a few lines of code. In addition, the package provides functionality for conventional approximate Bayesian computation, to compute model diagnostics, and to automatically estimate summary statistics. By virtue of being entirely written in JAX, \texttt{sbijax} is extremely computationally efficient, allowing rapid training of neural networks and executing code automatically in parallel on both CPU and GPU.    
\end{abstract}

\section{Introduction} 
\label{sec:intro}
Modern approaches to neural simulation-based inference (SBI) utilize recent developments in neural density estimation or score-based generative modelling to build surrogate models to approximate Bayesian posterior distributions. Similarly to conventional methods, such as approximate Bayesian computation (ABC) and its sequential (SMC-ABC) and annealing-based (e.g., SABC) variants, neural SBI methods infer this posterior distribution by first simulating synthetic data and then numerically constructing an appropriate approximation to this pseudo data set. SBI methods are attractive for a couple of reasons. On the one hand, this family of methods has been shown to be more computationally efficient and often more accurate than ABC methods, in particular for smaller simulation budgets. On the other hand, SBI allows to easily amortize inference, i.e., to infer the posterior distribution for multiple different observations once a neural model has been trained.

While a plethora of different models has been proposed in the recent literature, the development of adequate software packages has not followed at the same pace, and only few packages exist that allow modelers to use these methods. Most prominently, the {Python} package \texttt{sbi} \citep{tejero-cantero2020sbi} implements several approaches for neural simulation-based inference, such as a neural posterior, likelihood-ratio, and likelihood estimation \citep{cranmer2020frontier} utilizing a \texttt{PyTorch} backend \citep{paszke2019pytorch}. The package additionally provides an API for model diagnostics, such as posterior predictive checks, effective sample size computations and simulation-based calibration. However, the package lacks implementations of recent developments which pose the state-of-the-art in the field, such as by \citet{chen2023learning}, \citet{dirmeier2025simulationbased} or \citet{schmitt2023consistency}. Also, by virtue of being developed in \texttt{PyTorch} it is potentially restrictive to practitioners that do not have experience with it. For approximate Bayesian computation, several {Python} and {R} packages are available \citep{schaelte2022,dutta2021abcpy,approxbayescomp}. In particular \texttt{abcpy} \citep{dutta2021abcpy} implements a multitude of different ABC algorithms. However, none of these packages implement modern (neural) SBI methods. Furthermore, loosely related to modern simulation-based inference, the {R} package \texttt{Infusion} \citep{rousset2017summary} provides functionality to approximate summary likelihood surfaces using density estimation with Gaussian mixtures, but cannot be used for neural SBI.

Here we propose \texttt{sbijax}, a {Python} package implementing state-of-the-art methodology of neural simulation-based inference. While the main focus of the package is the implementation of recent algorithms to make them available to practitioners, e.g., \citet{wildberger2023flow} or \citet{schmitt2023consistency}, \texttt{sbijax} also implements common methods from approximate Bayesian computation, e.g., SMC-ABC \citep{beaumont2009adaptive}, to have the entire SBI toolbox in one efficient package. In addition, \texttt{sbijax} provides functionality for model diagnostics, posterior visualization and Markov Chain Monte Carlo (MCMC) sampling. The package uses the high-performance computing framework \texttt{JAX} as a backend \citep{jax2018github}. Using \texttt{JAX} has several advantages, including a) that it uses the same syntax as \texttt{numpy} \citep{harris2020array} which enables a seamless transition for applied scientists who already are familiar with it, and b) that empirical evaluations have shown that \texttt{JAX} can be significantly faster that \texttt{PyTorch} (see, e.g., \citet{phan2019composable}).

The structure of the manuscript is as follows. Section~\ref{sec:background} reviews required methodological background on neural simulation-based inference and ABC, and illustrates the different families of methods, each of which are implemented in \texttt{sbijax}, with an example from the literature. Section~\ref{sec:package} introduces the functionality of the \texttt{sbijax} {Python} package. Section~\ref{sec:examples} illustrates the usage of \texttt{sbijax} on a challenging Bayesian model using several of the implemented algorithms and compares their performance. Section~\ref{sec:real-data} demonstrates \texttt{sbijax} using a data set of experimental electroencephalogram measurements. Summarizing thoughts and future developments are presented in Section~\ref{sec:summary}.

\section{Methodological background}
\label{sec:background}
We are interested in inferring the Bayesian posterior distribution
\begin{equation}
    \pi(\theta | y_{\text{obs}}) = \frac{1}{Z} \pi(y_{\text{obs}} | \theta) \pi(\theta)
\end{equation}
where the parameters $\theta$ are the quantities of interest, $y_{\text{obs}}$ is a single or a set of observations, and $Z = \int \pi(y_{\text{obs}} | \theta) \pi(\theta) \mathrm{d}\theta$ is a normalization constant that does not depend on $\theta$ and, in Bayesian statistics, it is also referred to as the marginal likelihood.

In simulation-based inference (SBI) and approximate Bayesian computation (ABC), the likelihood function $\pi(y|\theta)$ is assumed to be inaccessible, i.e., intractable to compute, since it, e.g., involves solving high-dimensional integrals. For instance, consider the following example:
\begin{equation}
\begin{split}
f(y) &= \tfrac{1}{2} [1 + \text{erf}( \tfrac{y  - b_1}{w_1})] [1 - \text{erf} (\tfrac{y  - b_2}{w_2} ) ] \\ 
\theta &\sim G \\
 \alpha_t &\sim \mathcal{U}(\theta_1, \theta_1 + \theta_2), \quad \epsilon_t \sim \mathcal{U}(0, \theta_3), \\
y_{t + 1} &= \alpha_t f(y_t) y_t  + \epsilon_t
\end{split}
\label{eqn:solardynamo}
\end{equation}
where $b_1, w_1, b_2, w_2$ are constants, $\text{erf}$ is the error function, $G(\theta)$ is some prior distribution, $\mathcal{U}(a, b)$ is a uniform distribution bounded on the interval $[a, b]$, and $\theta = [\theta_1, \theta_2, \theta_3]^T$ is the vector of random parameters for which the posterior distribution should be inferred. Models of this kind have been used to simulate the variable magnitude of solar cycles \citep{charbonneau2005fluctuations}. The structural equation producing $y_{t+1}$ contains two non-Gaussian error terms, $\alpha_t$ and $\epsilon_t$, yielding a likelihood function that is outside the exponential family and that is difficult to derive analytically.

It is, however, possible to repeatedly simulate data from the prior predictive distribution, i.e., sample from the generative model given a realization from the prior and by that build a data set of synthetic observations. For instance, in equation~\eqref{eqn:solardynamo}, one can simulate $N$ pairs $\{(y_n, \theta_n)\}_{n=1}^N$ by first sampling $\theta_n \sim \pi(\theta)$ and then $y_n \sim \pi(y | \theta_n)$, where (we slightly abuse notation denoting) $y_n =(y_{n1}, \dots, y_{nT})$ and $T$ is a constant denoting the number of time steps to be simulated. In the following, will refer to this sampling procedure as \textit{simulator function} or \textit{observation model}, and denote the function that simulates realizations of the data and prior as \textit{generative model}.

Similarly to ABC \citep{sisson2018handbook}, neural simulation-based inference algorithms aim to infer the Bayesian posterior distribution
$\pi(\theta | y_{\text{obs}})$ for a parameter $\theta$ given a measurement $y_{\text{obs}}$ by utilizing synthetic data $\{(y_n, \theta_n)\}_{i=1}^N$ that have been simulated from a generative model for which the evaluation of the likelihood function is intractable. Specifically, this family of methods takes the simulated data set and then finds an approximation to the posterior using different approaches that estimate a neural-network parameterized model $q_\phi(y, \theta)$ with trainable parameters $\phi$ (which we will occasionally omit for ease of notation). In the following, we will refer to the family of ABC methods and {neural} SBI methods as \textit{simulation-based inference}. SBI methods can generally  be divided into four classes which are methods for
\begin{itemize}
    \item neural likelihood estimation (NLE) which estimate $q_\phi(y, \theta) \approx \pi(y|\theta)$,
    \item neural posterior estimation (NPE) which estimate $q_\phi(y, \theta) \approx \pi(\theta | y)$,
     \item neural likelihood-ratio estimation (NRE) which estimate $q_\phi(y, \theta)\approx \frac{\pi(y | \theta)}{\pi(y)}$,
    \item or approximate Bayesian computation (ABC) which is a distinct family that estimates a posterior distribution by comparing statistics of simulated data and observed data. 
\end{itemize}
In the following sections, we give an overview of the four classes of SBI. Additionally, for each class we introduce a concrete algorithm from the recent literature to better illustrate the respective approach to SBI.

\subsection{Neural likelihood estimation}
\label{sec:background-nle}

In order to infer the posterior distribution $\pi(\theta | y_{\text{obs}})$ given a specific observation $y_{\text{obs}}$, methods for neural likelihood estimation (NLE; e.g, \citet{papamakarios2019sequential,dirmeier2025simulationbased}) aim to approximate the intractable likelihood function with a surrogate model
\begin{equation}
q(y, \theta) \approx \pi(y|\theta)
\end{equation}
trained using simulated data pairs $\{(y_n, \theta_n)\}_{i=1}^N$. Having a trained surrogate, NLE methods construct an approximate posterior
\begin{equation}
\hat{\pi}(\theta | y_{\text{obs}} ) \propto \hat{q}(y_{\text{obs}}, \theta) \pi(\theta).
\label{eqn:nle-posterior}
\end{equation}
The surrogate model for the likelihood function ${q}(y, \theta)$ can generally be any method that allows to compute conditional density estimates, but is typically chosen to be a mixture model, a mixture density network \citep{bishop1994mixture}, or a conditional normalizing flow \citep{papamakarios2021normalizing}. 

Having access to the unnormalized posterior $\hat{q}(y_{\text{obs}}, \theta) \pi(\theta)$, it is possible to draw samples either using Markov Chain Monte Carlo methods (MCMC; \citet{tierney1994markov,compbayes2019}) or by approximating the posterior variationally \citep{blei2017variational}.

\subsubsection{Flow-based likelihood estimation}

Normalizing flow-based density estimators have recently been successfully applied for inference in various complex generative models \citep{papamakarios2019sequential}. Normalizing flows define pushforward measures that are parameterized by neural networks:
\begin{equation}
\hat{q}(y, \theta) = M_{\Psi} [ \pi_\text{base}(\cdot) ](y)
\end{equation}
where $\pi_\text{base}(z)$ is a base measure and $M$ denotes a pushforward operator. The operator maps the density $\pi_\text{base}(z)$ to $q_{\phi}(y, \theta)$ using a bijective deterministic transform $\Psi := \Psi_\phi(z; \theta)$ that is conditioned on a parameter sample $\theta$ and parameterized by a neural network with weights $\phi$, such that
\begin{equation}
\begin{split}
z \sim \pi_\text{base}(z) \\
y = \Psi(z; \theta).
\end{split}
\end{equation}
Thanks to the bijectivity constraint which is required to be able to both sample random variables and evaluate the probability density for a variable, the conditional transform $y = \Psi(z; \theta)$ has an inverse function $z = \Psi^{-1}(y; \theta)$.

Flow-based NLE methods first simulate a synthetic data set $\{(y_n, \theta_n)\}_{n=1}^N$ and fit the density estimator to it by optimizing the maximum likelihood objective
\begin{equation}
\begin{split}
\hat{\phi} 
&= \arg \max_\phi \mathbb{E}_{\theta, y}\left[ \log q(y, \theta) \right]\\
&= \arg \max_\phi \mathbb{E}_{\theta, y}\left[\log \pi_{\text{base}}(\Psi^{-1}(y; \theta))  + \log \bigg| \det \frac{\partial \Psi^{-1}}{\partial y} \bigg| \right]
\end{split}
\end{equation}
where the expectation is taken over $\theta \sim \pi(\theta)$, $y \sim \pi(y | \theta)$, and where $\det \frac{\partial \Psi^{-1}}{\partial y}$ is the determinant of the Jacobian matrix of $\Psi^{-1}$ which is obtained by the change-of-variables formula for continuous variables.

Having trained the neural network model, a surrogate model for the true posterior distribution is constructed as in equation~\eqref{eqn:nle-posterior}, i.e., $\hat{\pi}(\theta|y_{\text{obs}}) \propto \hat{q}(y_{\text{obs}}, \theta) \pi(\theta)$ which can be sampled from using MCMC methods.

\subsection{Neural posterior estimation}
\label{sec:background-npe}

Methods for neural posterior estimation (NPE; e.g.,\citet{greenberg2019automatic,wildberger2023flow,schmitt2023consistency}) approximate the posterior distribution $\pi(\theta | y_{\text{obs}})$ directly by learning a surrogate model of the form:
\begin{equation}
q(y, \theta) \approx \pi(\theta | y)
\end{equation}
which, similarly to Section~\ref{sec:background-nle}, is trained on synthetic data $\{(y_n, \theta_n)\}_{n=1}^N$ sampled from the generative model. NPE methods consequently construct the approximate posterior for an observation $y_{\text{obs}}$ directly using the trained model via
\begin{equation}
\hat{\pi}( \theta | y_{\text{obs}} ) = \hat{q}(y_{\text{obs}}, \theta)
\end{equation}

As for NLE, NPE methods utilize density estimators such as mixture density networks or normalizing flows to approximate the posterior distribution. However, the roles of observations $y$ and parameters $\theta$ are switched meaning that a density model is learned for the parameters and not the observations. 

The advantage of this family of methods is that, in general, no costly MCMC sampler has to be run to sample from the posterior. Rather, e.g., in the case of flow-based NPE methods, a sample from the base distribution can be drawn and then deterministically transformed forward to yield a sample from the approximate posterior. NPE methods, however, suffer from the fact that the produced posterior samples do not necessarily respect the constraints of the prior. For instance, the domain of a prior with box-constraints $\text{dom}(\pi(\theta))$, will possibly not be respected by the trained neural network, i.e., $\text{dom}(\pi(\theta)) \ne \text{dom}(q(y, \theta))$ unless additional {constraining bijections} are added to the model. 

\subsubsection{Flow matching posterior estimation}

As an example NPE model, we illustrate \textit{flow matching posterior estimation} (FMPE; \citet{wildberger2023flow}). FMPE proposes to use flow matching to train a continuous normalizing flow (CNF; \citet{chen2018neurao,liu2023flow}) as a surrogate for the posterior distribution \citep{lipman2023flow}. Similarly to conventional normalizing flows, CNFs define a pushforward measure but unlike them,  they use a time-dependent map $\Lambda_t(\theta) := \Lambda_{t}(\theta; y)$, with $t\in [0, 1]$, parameterized as an ordinary differential equation (ODE) instead of a discrete mapping:
\begin{equation}
\begin{split}
    \frac{\mathrm{d}}{\mathrm{d}t} \Lambda_t(\theta) &= v_{t}(\Lambda_t(\theta))\\
    \Lambda_{0}(\theta) &= \theta
\end{split}
\label{eqn:ode-npe}
\end{equation}
In equation~\eqref{eqn:ode-npe}, $v_{\phi}: [0, 1] \times \mathbb{R}^{d_\theta + d_y} \rightarrow \mathbb{R}^{d_\theta}$ is a vector field that is modelled with a neural network with trainable weights $\phi$. Interestingly, and in contrast to conventional normalizing flows, the transform $v_t$ can be specified by any neural network and does not need to be bijective by design.

The density defined by the pushforward is given by
\begin{equation}
\begin{split}
q_1(y, \theta) 
 &= T_{\Lambda_1} [ \pi_\text{base}(\cdot)](\theta) \\
&= q_0(\theta) \exp \left(  -\int_0^1 \text{div}\ v_{t}(\theta_t) \mathrm{d}t  \right)
\end{split}
\label{eqn:cnf}
\end{equation}
where we slightly abused notation and denote $q_0(\theta_0) = \pi_{\text{base}}$ as the base distribution that is used to sample the initial $\theta_0 \sim q_0$.  As before, the flow is trained using maximum likelihood:
\begin{equation*}
\hat{\phi} = \arg \max_\phi \mathbb{E}_{\theta, y} \left[ \log q_1(y, \theta) \right]
\end{equation*}
where the expectation is again taken over $\theta \sim \pi(\theta)$ and $y \sim \pi(y | \theta)$. Since the maximum likelihood objective above requires many network passes to solve the ODE in Equation~\eqref{eqn:ode-npe}, training the flow is in practice often not feasible. Instead, \citet{lipman2023flow} propose an alternative training objective which is computationally more favourable and which directly regresses $v_t$ on a new vector field $u_t$. Starting from a user-specified base distribution $\varrho_0(\theta)$, e.g., a multivariate Gaussian with diagonal covariance matrix, the vector field $u_t$ induces a probability path $\varrho: [0, 1] \times \mathbb{R}^{d_\theta} \rightarrow  \mathbb{R}^+$, such that $\int \varrho_t(\theta) \mathrm{d}\theta = 1$. The key insight of \citet{lipman2023flow} is that if the probability path and vector field are instead chosen to be conditional distributions i.e., $\varrho_t(\theta_t | \theta_1)$ and $u_t(\theta_t |\theta_1)$ such as
\begin{equation}
\begin{split}
\varrho_0(\theta |\theta_1) &= \mathcal{N}(\theta; 0, I) \\
\varrho_1(\theta |\theta_1) &= \mathcal{N}(\theta; \theta_1, \sigma^2I)  \\
\end{split}
\end{equation}
then the training objective can be framed as a MSE loss.
 \citet{lipman2023flow} discuss several possibilities to define the probability paths $\varrho_t$ and vector fields $u_t$ and propose, amongst others, to use the ones that are defined by the optimal transport map:
\begin{equation}
\begin{split}
\varrho_t(\theta | \theta_1) &= \mathcal{N}(t\theta_1, (1 - (1 - \sigma_\text{min})t)^2 I) \\
u_t(\theta | \theta_1) &= \frac{\theta_1 - (1 -  \sigma_\text{min})\theta}{1 - (1 - \sigma_\text{min})\theta}
\end{split}
\end{equation}
where $\sigma_\text{min}$ is a hyperparameter. To train the FMPE model, \citet{wildberger2023flow} suggest finding the parameters that minimize the following least squares loss
\begin{equation}
\hat{\phi} = \arg \min_{\phi} \mathbb{E}_{t,\theta_1,y,\theta_t}||v_t(\theta_t) - u_t(\theta_t|\theta_1)||^2
\end{equation}
where the expectation is over $t \sim \mathcal{U}(0, 1)$, $\theta_1 \sim \pi(\theta)$, $y \sim \pi(y | \theta_1)$, $\theta_t \sim \varrho_t(\theta_t | \theta_1)$.

Having a trained neural network $v_t$, the FMPE posterior is defined as in equation~\eqref{eqn:cnf}, i.e., $\hat{\pi}(\theta|y_{\text{obs}}) = q_1(y_{\text{obs}}, \theta)$, which can be computed with conventional ODE-solvers.

\subsection{Neural likelihood-ratio estimation}
\label{sec:background-nre}

Neural likelihood-ratio estimation methods (NLR, e.g., \citet{miller2022contrastive}) aim to approximate the intractable likelihood-to-evidence ratio 
\begin{equation}
    q(y, \theta) \approx r(y, \theta) = \frac{\pi(y|\theta)}{\pi(y)} = \frac{\pi(y, \theta)}{\pi(y) \pi(\theta)}
    \label{eqn:lr-new}
\end{equation}
trained on synthetic data and to use it to construct a surrogate posterior approximation
\begin{equation}
    \hat{\pi}(\theta | y_{\text{obs}}) = \hat{q}(y_{\text{obs}}, \theta)\pi(\theta)
\end{equation}
The denominator on the right-hand side of equation~\eqref{eqn:lr-new} does not depend on the parameters $\theta$ and the entire expression can consequently be approximated using Monte Carlo methods.

In the context of SBI, an approach to fit density-ratios consists in training a binary classifier $\varpi_\phi: \mathcal{Y} \times \Theta \rightarrow [0, 1]$ with trainable parameters $\phi$ that distinguishes samples that are drawn from the joint $\pi(y|\theta) \pi(\theta)$ from samples that that are drawn from the marginals $\pi(y) \pi(\theta)$ \citep{sugiyama2012density,hermans2020likelihood}. Specifically, to cast the density-ratio problem as a binary classification problem, the following conditional distribution for the data and parameters is assumed 
\begin{equation}
    \pi(y, \theta | c) := \begin{cases}
    \pi(y) \pi(\theta),              &\text{if } c=0\\
    \pi(y|\theta) \pi(\theta),& \text{if } c=1\\
\end{cases}
\end{equation}
where the class labels are assumed to be uniformly distributed, i.e., $\pi(c) = \tfrac{1}{2}$. For a cross-entropy loss, the Bayesian optimal classifier has the form
\begin{equation}
    \varpi(y, \theta) = \frac{\pi(y, \theta)}{\pi(y, \theta) + \pi(y)\pi(\theta)} = {sigm} \left( \log \frac{ \pi(y, \theta) }{ \pi(y) \pi(\theta) } \right)
\end{equation}
where $sigm$ is the sigmoid function. The Bayes optimal classifier approximates the conditional $\pi(c=1|y, \theta)$. 

An estimator for likelihood-to-evidence ratio can then be constructed by training a neural network such that
\begin{equation}
    \log \hat{q}(y, \theta) = \log \hat{r}(y,\theta) = \text{logit}\left( \hat{\varpi}(y, \theta) \right)
\end{equation}
where $\hat{\varpi}$ is the trained classifier. With the trained classifier, the approximate posterior using likelihood-ratio estimation is defined as in equation~\eqref{eqn:lr-new}. However, since the classifier cannot be generally trained to optimality, the approximate posterior is typically constructed as 
\begin{equation}
    \hat{\pi}(\theta | y_{\text{obs}}) = \frac{\hat{q}(y_{\text{obs}}, \theta)\pi(\theta)}{\int \hat{q}(y_{\text{obs}}, \theta)\pi(\theta) \mathrm{d} \theta}
\end{equation}
and sampled using MCMC methods.

\subsubsection{Contrastive neural ratio estimation}

Recently, \citet{miller2022contrastive} proposed an approach, called \textit{contrastive neural ratio estimation}, which instead proposes to use a multi-class classifier. We illustrate the approach below. The proposed method introduces a classifier aimed at detecting which, if any, of $C$ parameter sets $\theta_1, \cdots, \theta_C$, was responsible for generating a specific observation $y$. The assumed conditional distribution is given by
\begin{equation}
    \pi(y, \theta|c) = \begin{cases}
    \pi(y) \pi(\theta_1)\cdots\pi(\theta_C)& \text{if } c=0\\
    \pi(y|\theta_c) \pi(\theta_1)\cdots\pi(\theta_C),              &\text{if } c=1, \cdots, C
\end{cases}
\end{equation}
\citet{miller2022contrastive} define the class probabilities, $p(c) = p_C$ for all $ c > 0$ and $p(c=0) = p_0 = 1 - Cp_C$. This defines the odds of a pair $(y, \theta)$ being drawn dependently versus independently to be $\gamma = \frac{Cp_c}{1 - Cp_c}$. Given these definitions, the conditional class probabilities are given by:
\begin{equation}
    \pi(c|y, \theta_1, \dots, \theta_C) = \begin{cases}
    \frac{K}{K + \gamma \sum_{i=1}^C r(y, \theta_i) }& \text{if } c=0\\[2.2ex]
    \frac{\gamma  r(y, \theta_c) }{K + \gamma \sum_{i=1}^C r(y , \theta_i)},              &\text{if } c=1, \cdots, C
\end{cases}
\end{equation}
where $r(y , \theta) = \frac{\pi(y|\theta)\pi(\theta)}{\pi(y)\pi(\theta)}$ is the likelihood-to-evidence ratio. The conditional probability function can, as before, be approximated with a classifier $\varpi_\phi: \mathcal{Y} \times \Theta \rightarrow [0, 1]$ which uses a neural network $h_\phi:\mathcal{Y} \times \Theta \rightarrow \mathbb{R}^+$ with trainable parameters $\phi$:
\begin{equation}
    \varpi(c|y, \theta_1, \dots, \theta_C) := \begin{cases}
    \frac{K}{K + \gamma \sum_{i=1}^C h(y, \theta_i) }& \text{if } c=0\\[2.2ex]
    \frac{\gamma h(y |\theta_c) }{K + \gamma \sum_{i=1}^C h(y,\theta_i)},              &\text{if } c=1, \cdots, C
\end{cases}
\end{equation}

To encourage the neural network to converge to the likelihood-ratio, i.e., $h(y, \theta) \approx r(y, \theta)$, \citet{miller2022contrastive} train the classifier with a cross-entropy loss such that
\begin{equation*}
\begin{split}
\hat{\phi} &= \arg \max_{\phi} \mathbb{E}_{\theta,y,c}\left[\log \varpi (c|y, \theta_1, \cdots, \theta_C)\right] \\ 
&= \arg \max_{\phi} p_0 \mathbb{E}_{\theta,y|c=0}\left[\log \varpi (c=0|y, \theta_1, \cdots, \theta_C)\right] +
p_C \sum_{k=1}^C\mathbb{E}_{\theta,y|c=k}\left[\log \varpi (c=k|y, \theta_1, \cdots, \theta_C)\right]
\\ 
&= \arg \max_{\phi} p_0 \mathbb{E}_{\theta,y|c=0}\left[\log \varpi (c=0|y, \theta_1, \cdots, \theta_C)\right] +
C p_C \mathbb{E}_{\theta,y|c=C}\left[\log \varpi (c=C|y, \theta_1, \cdots, \theta_C)\right]
\end{split}
\end{equation*}

After training, the approximate posterior is constructed via
\begin{equation*}
    \hat{\pi}(\theta | y_{\text{obs}}) \propto \hat{h}(y_{\text{obs}}, \theta) \pi(\theta)
\end{equation*}
from which samples can be drawn using MCMC methods.

\subsection{Approximate Bayesian computation}
\label{sec:background-abc}

In comparison to neural SBI methods, approximate Bayesian computation using rejection sampling (Rejection ABC; \citet{sisson2018handbook}), sequential variants thereof (SMC-ABC; e.g., \citet{beaumont2009adaptive}) and simulated-annealing ABC (SABC; e.g., \citet{albert2015simulated}) aim to infer the posterior distribution $\pi(\theta | y_{\text{obs}})$ by direct comparison of simulated synthetic data to observed data. 

Specifically, vanilla ABC first simulates a measurement-parameter pair $(y, \theta)$ from the prior predictive distribution and then accepts the prior sample $\theta$  if the distance between simulated and observed data, $y$ and $y_\text{obs}$, respectively, is sufficiently small, i.e., if the distance $d(y, y_\text{obs}) < \epsilon$ is smaller than some threshold $\epsilon$ (where the distance measure $d$ is user defined). The described procedure is equivalent to drawing a sample $(y, \theta)$ from the distribution proportional to
\begin{equation*}
    I\left( d(y, y_\text{obs}) < \epsilon \right) \pi(y|\theta)\pi(\theta)
\end{equation*}
which when marginalizing out the synthetic data $y$ recovers the joint distribution $\pi(y_{\text{obs}}|\theta)\pi(\theta)$ in the limit $\epsilon \rightarrow 0$:
\begin{equation*}
     \lim_{\epsilon \rightarrow 0} \int I\left( d(y, y_\text{obs}) < \epsilon \right) \pi(y|\theta)\pi(\theta) \mathrm{d} y = \pi(y_{\text{obs}}|\theta)\pi(\theta)
\end{equation*}
As a consequence, the ABC procedure introduced above, draws samples from the exact posterior $\pi(\theta|y_{\text{obs}})$ if and only if $\epsilon \rightarrow 0$.

This derivation of the ABC posterior has several drawbacks. Using a discrete indicator function $I$ does not allow to distinguish samples $\theta, \theta'$ for which the induced distances are very different, i.e., $d(g(\theta), y_{\text{obs}}) \ll d(g(\theta'), y_{\text{obs}})$, which leads to a general loss of information about $\theta$. As a remedy, one can introduce a \textit{smooth} semi-definite function $K_\epsilon$ to replace the indicator function $I$. We can, for instance, choose $K_\epsilon$ as a kernel 
\begin{equation*}
    K_\epsilon({d}(y, y')) =  K_\epsilon(u) = \frac{1}{\epsilon} K \left( \frac{u}{\epsilon}  \right)
\end{equation*}
where we use the conventional definition of a kernel function (i.e., $K(u) \ge 0$, $\int K(u)\mathrm{d}u = 1$, $\int uK(u)\mathrm{d}u = 0$ and $\lim_{\epsilon \rightarrow 0} K_\epsilon(0) = 1$ is a Dirac delta function).

The new ABC posterior is constructed as before, replacing the indicator with the kernel
\begin{equation*}
    \hat{\pi}(\theta | y_{\text{obs}}) \propto \int K_\epsilon \left( d(y, y_\text{obs}) \right) \pi(y|\theta)\pi(\theta) \mathrm{d} y
\end{equation*}
For $\epsilon \rightarrow 0$ the kernel converges to a Dirac delta function $K_\epsilon \left( d(y,  y_{\text{obs}}) \right) \rightarrow \delta(d(y,  y_{\text{obs}}))$, with $\delta(0) = 1$  and $\delta(u) = 0$ otherwise, and the ABC posterior approximation recovers the true posterior $\pi(\theta|y_{\text{obs}})$. 

This construction is still inefficient, since for continuous simulated $y$ and for $\epsilon \rightarrow 0$ the distance $d(y, y_\text{obs})$ is always greater than zero which means that $K_\epsilon(y, y_{\text{obs}}) \approx 0$ and that each sample $\theta$ gets discarded. In practice, ABC methods consequently define $\epsilon > 0$ and accept a sample $\theta \sim \pi(\theta)$ with a Metropolis acceptance probab proportional to $K_\epsilon(d(y, y_{\text{obs}}))$.

Furthermore, in practice, the ABC posterior above is rarely used, since it is only possible to simulate data $y$, such that $y \approx y_{\text{obs}}$ for very low-dimensional problems. Consequently, first the data is usually reduced to a set of {summary statistics} $s = S(y)$, where $S$ is a function that computes the summaries, and then a distance between the summaries is computed. A prior sample $\theta$ is then accepted in the set of posterior samples with a probability proportional to $K_\epsilon(s, s_{\text{obs}})$.
For large and complex data instead of comparing $y$ and $y_{obs}$, the comparison is performed using a summary statistics $S$
The ABC approximation to $\pi(\theta | s_{\text{{obs}}})$ for a specific summary statistics of the observed data, $S(y_{\text{obs}})$, is given by
\begin{equation*}
    \hat{\pi}(\theta | s_{\text{obs}}) \propto \int K_\epsilon \left( d(s,  s_{\text{obs}}) \right) \pi(s|\theta) \pi(\theta) \mathrm{d}s
\end{equation*}
where $K_\epsilon$ is again a kernel function, and $\pi(s| \theta)$ denotes the (intractable) likelihood function of the summary $s$ implied by $\pi(y|\theta)$ and $S$. 

As before, for $\epsilon \rightarrow 0$ the ABC posterior approximation recovers the true posterior $\pi(\theta|s_{\text{obs}})$. If and only if, the summary $S$ is {sufficient}, no additional layer of approximation is added.

\subsubsection{Sequential Monte Carlo ABC}
The SMC-ABC implementation in \texttt{sbijax} is a modified version of the ABC samplers by \citet{beaumont2009adaptive} and \citet{del2012adaptive} (c.f. Algorithm~4.8 in \citet{sisson2018handbook}). SMC-ABC methods make use of a similar approach as  rejection ABC methods with the same theoretical guarantees as outlined above. In addition, they implement resampling and MCMC transition steps, which typically improves the quality of the approximated posterior distribution. SMC-ABC proceeds in $R$ rounds, where, in each round $r$, a particle $\theta^r_n \sim \pi(\theta)$ is perturbed using a MCMC transition kernel $\pi(\theta |\theta^{r - 1}_n)$. Additionally, to improve the convergence to the true posterior distribution the  distance threshold $\epsilon^r$  used by the kernel function $K_{\epsilon^r}$ is reduced in each round, and the set of all particles is resampled to maintain a high effective sample size, i.e., to discard particles from regions with low density. Self-explanatory pseudo-code of our implementation is given in algorithm~\ref{alg:smcabc}.

\RestyleAlgo{ruled} 
\SetKwComment{Comment}{/* }{ */}
\renewcommand{\KwData}[1]{\textbf{Inputs: }#1\\}
\renewcommand{\KwResult}[1]{\textbf{Outputs: }#1\\}
\begin{algorithm}
\caption{SMC-ABC (as implemented in \texttt{sbijax})}
\label{alg:smcabc}
\KwData{Observation $y_{\text{obs}}$, prior distribution $\pi(\theta)$, simulator function $\pi(y| \theta)$, transition kernel $\pi(\theta|\theta')$,
distance function $d$, summary statistic function $S$, particle size $N$, number of rounds $R$, epsilon decay rate $\gamma_\epsilon$}
\KwResult{Particles $\{\theta_n^R\}_{n=1}^N$ as approximate posterior distribution}
\textit{Initialize:}\\
Compute $s_{\text{obs}} = S(y_\text{obs})$\\
\For{$n \gets 1$ \KwTo $N$}{
Sample $\theta_n^0 \sim \pi(\theta)$\\
Simulate $y_n^0 \sim \pi(y | \theta_n^0)$\\ 
Compute summary statistics $s_n^0 = S(y_n^0)$ \\
Set weights $w_n^0 = \tfrac{1}{N}$
}
Set $\epsilon^1 = \min_n d(s_n^0, s_{\text{obs}})$\\
\textit{Sample:}\\
\For{$r \gets 1$ \KwTo $R$}{
\For{$n \gets 1$ \KwTo $N$}{
Sample $\theta_n^r \sim \pi(\theta_n^r|\theta_n^{r-1})$ and $y_n^r \sim \pi(y|\theta_n^r)$ until $d\left( S(y_n^r), s_{\text{obs}} \right) < \epsilon^r$\\
Compute summary statistics $s_n^r = S(y_n^r)$ \\
Set weights $w_n^r \propto \frac{\pi(\theta_n^r)}{\sum_j w_m^{r-1} \pi(\theta_n^r|\theta_m^{r-1}) }$ \\
\If{the effective sample size is too low} {
Re-sample $\theta_n^r$ from the set of all $\theta_n^{r}$ with probabilities $w_n^{r}$ \\
Set weights $w_n^r = \tfrac{1}{N}$
}
}
Set $\epsilon^{r + 1} = \gamma_\epsilon\epsilon^{r}$\\
}
\end{algorithm}

\RestyleAlgo{ruled} 
\SetKwComment{Comment}{/* }{ */}
\renewcommand{\KwData}[1]{\textbf{Inputs: }#1\\}
\renewcommand{\KwResult}[1]{\textbf{Outputs: }#1\\}
\begin{algorithm}
\caption{Sequential neural SBI (as implemented in \texttt{sbijax})}
\label{alg:sequential-sbi}
\KwData{Observation $y_{\text{obs}}$, prior distribution $\pi(\theta)$, simulator function $\pi(y|\theta)$, trainable surrogate model $q(y, \theta)$, 
number of particles $N$, number of rounds $R$}
\KwResult{Posterior approximation $\hat{\pi}^R(\theta | y_{\text{obs}})$}
\textit{Initialize:}\\
Set $\hat{\pi}^0(\theta | y_{\text{obs}}) = \pi(\theta)$ and $\mathcal{D} = \{\}$\\
\textit{Iterate:}\\
\For{$r \gets 1$ \KwTo $R$}{
\For{$n \gets 1$ \KwTo $N$}{
Sample $\theta_n \sim \hat{\pi}^r(\theta | y_{\text{obs}})$ \\
Simulate $y_n \sim \pi(y | \theta_n)$\\
}
Concatenate $\mathcal{D} = \mathcal{D} \cup \{(y_n, \theta_n) \}_{n=1}^N$\\
Train $q(y, \theta)$ on $\mathcal{D}$ using methods described in \cref{sec:background-nle,sec:background-npe,sec:background-nre} and construct approximate posterior $\hat{\pi}^r(\theta|y_{\text{obs}})$ with it.
}
\end{algorithm}

\subsection{Sequential inference}
\label{sec:background-sequential-inference}

The neural SBI methods discussed in \cref{sec:background-nle,sec:background-npe,sec:background-nre} construct surrogate posterior distributions by computing approximations to the likelihood function, likelihood-to-evidence-ratio, or the posterior distribution directly. Since they are trained from simulated batches of data and can then be applied to any given observation, the approach is conventionally referred to as \textit{amortized inference} (c.f. section~\ref{sec:background-abc} where the posterior is necessarily computed for a specific observation $y_{\text{obs}}$).

Some neural simulation-based inference methods support sequential inference which has been shown empirically to improve the quality of the approximate posterior distributions. Similarly to SMC methods, sequential SBI approaches proceed in multiple rounds $r$:
\begin{itemize}
\item In round $r$, sample parameters $\theta^r_n \sim \hat{\pi}^{r-1}(\theta|y_{\text{obs}})$ and observations $y_n^r \sim \pi(y | \theta^r_n)$ where $\hat{\pi}^{r -1}(\theta|y_{\text{obs}})$ is the prior $\pi(\theta)$ if $r=1$ and is the trained surrogate posterior if $r > 1$.
\item Append data $\mathcal{D}_r = \{(y^r_n, \theta^r_n) \}_{n=1}^N$ to a data set $\mathcal{D}$ that consists of all samples
\item Construct a surrogate posterior $\hat{\pi}_r(\theta | y_{\text{obs}})$ given an observation $y_{\text{obs}}$ and data $\mathcal{D}$ using the methods from \Cref{sec:background-nle,sec:background-npe,sec:background-nre}
\end{itemize}

Methods that allow sequential inference necessarily condition on a specific measurement $y_0$ in each round which makes inferential accuracy of the posterior better, but has the disadvantage that the procedure is not amortized anymore, i.e., if the posterior $\pi(\theta|y_1)$ for a measurement $y_1$ should be computed, the algorithm needs to be re-trained. Since in typical applications, e.g., in biology or physics, the measurement is, however, generally fixed, this is not a major drawback. Algorithm~\ref{alg:sequential-sbi} delineates the procedure for general neural SBI methods.

\section{The {sbijax} package} 
\label{sec:package}

The {Python} package \texttt{sbijax} provides high-quality implementations of methods for neural posterior, neural likelihood and neural likelihood-ratio estimation, and approximate Bayesian computation. The package also contains methods to automatically learn summary statistics of high-dimensional data using neural networks which can then be used for further analysis, e.g., using NLE or ABC methods. In addition, \texttt{sbijax} offers functionality for simple construction of neural networks, computing model diagnostics and for visualization of posteriors and these diagnostics. \texttt{sbijax} is built on the high-performance computing library \texttt{JAX} in a computationally efficient and light-weight framework.

\begin{table}[h]
    \centering
    \caption{Implemented SBI methods in \texttt{sbijax}.}
    \begin{tabular}{lll}
        \toprule
        \textbf{Models} & {\bf Class name} & \textbf{Main reference}\\        
        \midrule
        Sequential Monte Carlo ABC & \texttt{SMCABC} & \citet{beaumont2009adaptive} \\
        Neural likelihood estimation & \texttt{NLE} & \citet{papamakarios2019sequential}        \\
        Surjective neural likelihood estimation & \texttt{SNLE} & \citet{dirmeier2025simulationbased} \\
        Automatic posterior transformation & \texttt{NPE} & \citet{greenberg2019automatic}\\
        Contrastive neural ratio estimation &\texttt{NRE} & \citet{miller2022contrastive}\\        
        Flow matching posterior estimation & \texttt{FMPE} & \citet{wildberger2023flow}\\
         Posterior Score Estimation                     & \texttt{NPSE}     & \citet{sharrock2024sequential}    \\
         All-In-One Posterior Estimation                & \texttt{AIO}      & \citet{gloeckler2024allinone}      \\
        Consistency model posterior estimation & \texttt{CMPE} & \citet{schmitt2023consistency}\\
        Neural approximate sufficient statistics & \texttt{NASS} & \citet{chen2021neural}\\
        Neural approximate slice sufficient statistics &\texttt{NASSS} & \citet{chen2023learning}\\                
        \bottomrule
    \end{tabular}
    \label{tab:all-methods}
\end{table}

The \texttt{sbijax} package implements a low-level object-oriented programming interface which follows similar tools in the \texttt{JAX}-verse \citep{jax2018github}, such as \texttt{Surjectors} \citep{dirmeier2024surjectors}, \texttt{Distrax} and \texttt{Haiku} \citep{deepmind2020jax}, and is fully compatible with each of them which allows seamless integration in statistics and machine learning projects that use \texttt{JAX}.

In the following sections, we give an overview of the general workflow for applying neural simulation-based inference in \texttt{sbijax} using a motivating example, i.e., the steps a user has to follow to compute a surrogate posterior model. The steps consist of
\begin{itemize}
    \item Mathematical definition of the prior model $\pi(\theta)$ and simulator function $\pi(y|\theta)$, and transcription as {Python} code,
    \item definition of a neural network and inferential algorithm,
    \item training and posterior inference,
    \item and visualization of posteriors and model diagnostics.
\end{itemize}
The section concludes with a demonstration of additional implemented functionality, i.e., sequential inference, automatic summary statistic computation, and available neural networks and MCMC samplers.

All inferential algorithms that are implemented in \texttt{sbijax} at the time of writing are shown in Table~\ref{tab:all-methods}, functionality to construct neural network architectures is given in Table~\ref{tabl:neural-networks}, and implemented MCMC samplers are shown in Table~\ref{tabl:samplers}. The entire documentation of the package with detailed instructions to use all algorithms and methods can be found online at \href{https://sbijax.rtfd.io}{\texttt{sbijax.rtfd.io}}. Experimental details can be found in Appendix~\ref{app:experimental-details}.

\subsection{Model definition}
\label{sec:sbijax-model_definition}

Construction of a model begins by defining the prior model $\pi(\theta)$ and the simulator function $g$. For instance, consider the following simple Gaussian model
\begin{equation}
\begin{split}
\mu &\sim \mathcal{N}_2(0, I)\\
\sigma &\sim \mathcal{N}^+(1)\\
y & \sim \mathcal{N}_2(\mu, \sigma^2 I)\\
\end{split}
\label{eqn:linear-gaussian}
\end{equation}
where $\mathcal{N}_2(0, I)$ denotes a bivariate normal distribution and $\mathcal{N}^+(1)$ a univariate half-normal distribution. In this example, the likelihood is simple to compute and conventional MCMC methods would be possible to use, but we will use it for for the sake of demonstration. The {Python} transcription of Equation~\eqref{eqn:linear-gaussian} has the following form:
\begin{small}
\begin{verbatim}
>>> import jax
>>> from jax import numpy as jnp, random as jr
>>> from tensorflow_probability.substrates.jax \ 
...     import distributions as tfd
>>>
>>> def prior_fn():
...     prior = tfd.JointDistributionNamed(dict(
...         mean=tfd.Normal(jnp.zeros(2), 1.0),
...         scale=tfd.HalfNormal(jnp.ones(1)),
...     ))
...     return prior
>>> 
>>> def simulator_fn(seed: jax.random.PRNGKey, theta: dict[str, jax.Array]):
...     noise = tfd.Normal(jnp.zeros_like(theta["mean"]), 1.0)
...     y = theta["mean"] + theta["scale"] * noise.sample(seed=seed)
...     return y
\end{verbatim}
\end{small}

The prior model is a \texttt{TensorFlow Probability} (TFP; \citet{dillon2017tensorflow}) object of class \texttt{JointDistributionNamed} that 
defines a joint prior over all latent variables, in this case for the mean and standard deviation of the observation model. The object allows to both sample from the joint prior as well as evaluate the probability of a sample. Using TFP distributions is convenient, because one can readily define highly complex prior models and internally TFP manages the computation of log-probabilities or drawing of random variables. In principle, the prior function can be any object that has access to two functions: a function called \texttt{sample} that returns random draws and takes as arguments a key for pseudo-random number generation, \texttt{seed}, and an integer tuple specifying the sample size, \texttt{sample\_shape},
\begin{small}
\begin{verbatim}
>>> prior = prior_fn()
>>> prior.sample(seed: jax.random.PRNGKey, sample_shape=())
\end{verbatim}
\end{small}
and a function \texttt{log\_prob} that accepts a list of random variables, \texttt{value}, and returns a list of log-probabilities:
\begin{small}
\begin{verbatim}
>>> prior.log_prob(value: jax.Array)
\end{verbatim}
\end{small}
The sample function of the prior, however, necessarily returns a dictionary where the keys are strings that specify the name of the parameter and the values are realizations of the marginal distributions.

The simulator function is a user-defined function that takes as function arguments a key of class \texttt{jax.random.PRNGKey} for pseudo-random number generation, called \texttt{seed}, and a hashmap of prior parameter values, \texttt{theta}, which can be generated by calling the prior function:
\begin{small}
\begin{verbatim}
>>> simulator_fn(seed: jax.random.PRNGKey, theta: dict[str, jax.Array])
\end{verbatim}
\end{small}
The simulator uses the arguments to generate a sample from the observation model.

\subsection{Algorithm definition}
\label{sec:sbijax-alg_definition}
Given the prior and simulator functions $\texttt{prior\_fn}$ and $\texttt{simulator\_fn}$, \texttt{sbijax} requires definition of an inferential algorithm and a trainable neural network. We will use neural likelihood estimation for the rest of this section, however, all other algorithms follow a similar workflow. The code excerpt below illustrates how objects for an inferential algorithm, here \texttt{NLE}, are constructed. 
\begin{small}
\begin{verbatim}
>>> n_dim_data = 2
>>> n_layers, hidden_sizes = 5, (64, 64)
>>> neural_network = make_maf(n_dim_data, n_layers=n_layers, hidden_sizes=hidden_sizes)
>>>
>>> fns = prior_fn, simulator_fn
>>> model = NLE(fns, neural_network)
\end{verbatim}
\end{small}
The first argument of \texttt{NLE}, and in fact all methods implemented in \texttt{sbijax}, is a tuple consisting of the prior model and the simulator function (see definition in Section~\ref{sec:sbijax-model_definition}). The second argument is the neural network that is to be used for learning a surrogate model. The allowed network architecture depends on the inferential algorithm. In the case of \texttt{NLE}, it is necessarily a conditional density estimator for the data $y$ since we approximate the likelihood function. Here, we use a normalizing flow normalizing flow with dimensionality $d_y=2$.

For all algorithms, we provide both high-level functionality to automatically construct a neural network for an algorithm and a low-level API which allows the user to fully customize a neural network model (see table~\ref{tabl:neural-networks} for an overview of pre-implemented neural networks). In the code excerpt above, we make use of the high-level functionality and define an affine masked autoregressive normalizing flow (MAF; \citet{papamakarios2017masked}) using the function
\begin{small}
\begin{verbatim}
>>> make_maf(n_dim: int, n_layers: int, hidden_sizes: Iterable[int])
\end{verbatim}
\end{small}
The arguments of the function specify the dimensionality of the modelled space (\texttt{n\_dim}), the number of flow layers to use (\texttt{n\_layers}) and the number of nodes per hidden layer of the neural network (\texttt{hidden\_sizes}).

To construct the same network as above using the low-level API, the following example code could be used:
\begin{small}
\begin{verbatim}
>>> def make_custom_affine_maf(n_dimension, n_layers, hidden_sizes):
...     def _bijector_fn(params):
...         means, log_scales = unstack(params, -1)
...         return ScalarAffine(means, jnp.exp(log_scales))
... 
...     def _flow(method, **kwargs):
...         layers = []
...         order = jnp.arange(n_dimension)
...         for _ in range(n_layers):
...             layer = MaskedAutoregressive(
...                 bijector_fn=_bijector_fn,
...                 conditioner=MADE(
...                     n_dimension, list(hidden_sizes), 2,
...                 ),
...             )
...             order = order[::-1]
...             layers.append(layer)
...             layers.append(Permutation(order, 1))
...         chain = Chain(layers)
...         base_distribution = tfd.Independent(
...             tfd.Normal(jnp.zeros(n_dimension), tfd.ones(n_dimension)),
...             reinterpreted_batch_ndims=1,
...         )
...         td = TransformedDistribution(base_distribution, chain)
...         return td(method, **kwargs)
...
...     td = hk.transform(_flow)
...     return td
>>>
>>> neural_network = make_custom_affine_maf(n_dim_data, n_layers, hidden_sizes)
>>> model = NLE(fns, neural_network)
\end{verbatim}
\end{small}
The low-level construction is aimed for users with advanced understanding of simulation-based inference and deep learning and who wish to be explicit in the neural architecture that they are using and exact details how the density estimators can be constructed can be found in the online documentation. To build the required neural network architectures, we use the \texttt{JAX}-based packages \texttt{Surjectors} \citep{dirmeier2024surjectors} and \texttt{Haiku} \citep{haiku2020github}.

\subsection{Training and Inference}
\label{sec:sbijax-training_and_inference}
Given an SBI object, in the above code example called \texttt{model}, training and posterior inference proceeds by first simulating data, then fitting the neural network, and finally sampling from the approximate posterior distribution. The code excerpt below illustrates the process:
\begin{small}
\begin{verbatim}
>>> data = model.simulate_data(seed=jr.PRNGKey(1), n_simulations=10_000)
>>> params, losses = model.fit(jr.PRNGKey(2), data=data)
>>> inference_results, diagnostics = model.sample_posterior(
...     jr.PRNGKey(3), params, y_obs
... )
\end{verbatim}
\end{small}
We describe the three steps in detail in the following paragraphs.

\subsubsection*{Data}

The function \texttt{model.simulate\_data} can be used to sample data from the simulator function and prior model and is exposed by every SBI algorithm. It has the following method declaration:
\begin{small}
\begin{verbatim}
>>> model.simulate_data(seed: jax.random.PRNGKey,  n_simulations: int)
\end{verbatim}
\end{small}
The function requires a random seed for pseudo-random number generation (\texttt{seed}) and an integer specifying the number of simulated points (\texttt{n\_simulations}) as mandatory arguments. Calling the function returns a \texttt{PyTree}, a dictionary of dictionaries, of pairs $\{(y_n, \theta_n)\}_{n=1}^N$:
\begin{small}
\begin{verbatim}
>>> print(data)
>>> {
...  'y': Array([[-0.01060869,  1.5589136 ],
...              [-0.592015  ,  1.6256292 ],
...               ...
...              [-2.3662274 ,  0.95785517]], dtype=float32), 
...  'theta': {
...         'mean': Array([[ 0.02413638,  1.6076895 ],
...                        [-0.35492337,  1.4311316 ],
...                         ...
...                        [-2.1717741 ,  1.2112944 ]], dtype=float32),
...         'scale': Array([[0.02382747],
...                         [0.16646351],
...                           ...
...                         [0.1992683 ]], dtype=float32), 
...     }
... }
\end{verbatim}
\end{small}
The object has one key-value pair that contains measurements, accessible with key \texttt{y}, and one key-value pair of parameters that is accessible with the key $\texttt{theta}$. Since the prior model $\pi(\theta)$ from Section~\ref{sec:sbijax-model_definition} contains a mean and scale parameter with names $\texttt{mean}$ and $\texttt{scale}$, the keys of the parameter dictionary have the same names as keys.

\subsubsection*{Training}
The generated data set can be used to fit the model using the function \texttt{model.fit} yielding a set of neural network weights, above called \texttt{params}, and a training and validation loss profile which can be used to validate that training converged successfully, called \texttt{losses}. The function has the following signature
\begin{small}
\begin{verbatim}
>>> model.fit(
...     seed: jax.random.PRNGKey, 
...     data: PyTree,
...     n_iter: int,
...     batch_size: int,
...     percentage_data_as_validation: float
... )
\end{verbatim}
\end{small}
The mandatory arguments of the functions are as following:
\begin{itemize}    
    \item \texttt{seed}: a \texttt{JAX} key for deterministic pseudo random number generation,
    \item \texttt{data}: a data set generated using \texttt{simulate\_data},    
    \item \texttt{n\_iter}: the maximal number of epochs to run the training optimizer,
    \item \texttt{batch\_size}: the size of each batch for which a gradient-update is computed,
    \item \texttt{percentage\_data\_as\_validation}: the percentage of the entire data set that is withheld during training and used as validation data set.
\end{itemize}

In comparison to object-oriented machine and deep learning tools such as \texttt{PyTorch} and \texttt{TensorFlow} where neural network weights are stored as member variables of an object, \texttt{sbijax} follows the functional programming paradigm of \texttt{JAX} and yields the set of network weights, \texttt{params}, as part of the return value of the function \texttt{fit}. Since \texttt{sbijax} uses the neural network library \texttt{Haiku} internally, the networks are merely dictionaries of arrays and matrices that can be easily serialized and stored on the hard drive. The loss profile, called \texttt{losses} above, can be used to check if the training of the neural network converged to a local minimum. The first column of that object represents losses on the training set, the second column represents losses of a withheld validation set that is used for early stopping
(see Figure~\ref{fig:bivariate_gaussian-losses} in the Appendix for a visualization thereof).

\subsubsection*{Posterior sampling}

Samples from the approximate posterior can be taken using the function \texttt{model.sample\_posterior}. The function has the following signature
\begin{small}
\begin{verbatim}
>>> model.sample_posterior(
...     seed: jax.random.PRNGKey, 
...     params: dict,
...     observable: int,
...     n_samples: int,
...     **kwargs
... )
\end{verbatim}
\end{small}
The function requires the neural network weights \texttt{params} that have been obtained using \texttt{model.fit} and an observation $y_\text{obs}$, called \texttt{observable}, in order to sample the approximate posterior $\hat{\pi}(\theta|y_\text{obs})$ using the trained surrogate model $\hat{q}(y, \theta)$. The argument \texttt{n\_samples} determines the size of the posterior sample. The {Python}-specific keyword-argument \texttt{**kwargs} determines additional hyper-parameters, for instance, the MCMC sampler to be used or number of MCMC chains to be sampled from in parallel, and is defined for each algorithm separately.

Calling \texttt{sample\_posterior} returns a tuple of two objects. The first element of the tuple, called \texttt{inference\_results} above, contains the posterior samples. The samples are wrapped within an object of the class \texttt{InferenceData} of the widely-used \texttt{Arviz} package for Bayesian computation \citep{kumar2019arviz}:
\begin{small}
\begin{verbatim}
>>> print(inference_results)
... Inference data with groups:
... 	> posterior
... 	> observed_data
\end{verbatim}
\end{small}
The object has two attributes called \texttt{posterior} and \texttt{observed\_data}. Accessing the posterior samples yields a data set of class \texttt{xarray} \citep{hoyer2017xarray} which is labelled multi-dimensional array and commonly used in data science with {Python}:
\begin{small}
\begin{verbatim}
>>> print(inference_results.posterior)
... <xarray.Dataset> Size: 280kB
... Dimensions:      (chain: 4, draw: 5000, mean_dim: 2, scale_dim: 1)
... Coordinates:
...   * chain        (chain) int64 32B 0 1 2 3
...   * draw         (draw) int64 40kB 0 1 2 3 ... 4997 4998 4999
...   * mean_dim     (mean_dim) int64 16B 0 1
...   * scale_dim    (scale_dim) int64 8B 0
... Data variables:
...     mean         (chain, draw, mean_dim) float32 160kB 0.7088 0.6792 ... -0.07558
...     scale        (chain, draw, scale_dim) float32 80kB 1.232 0.9525 ... 0.9773
\end{verbatim}
\end{small}

The second return value of $\texttt{sample\_posterior}$ is a dictionary of common (MCMC) model diagnostics which depends on the employed algorithm. When using methods that use MCMC methods to sample from the surrogate posterior, such as  \texttt{NLE} or \texttt{NRE}, the diagnostics consist of the normalized split-$\hat{R}$ \citep{vehtari2021split} as well as the (relative) effective sample size (ESS) for each parameter. For all other methods, only the ESS is returned.

\subsection{Model diagnostics and visualization}
The \texttt{sbijax} package provides basic functionality for computing model diagnostics, and posterior and model diagnostic visualization. All relevant functions return objects of the class \texttt{Axis} from the common \texttt{Matplotlib} plotting library which allows for user-defined customization, such as coloring, axis labels or font sizes \citep{hunter2007mat}.

To visualize posterior samples and MCMC traces, the functions \texttt{plot\_posterior} and \texttt{plot\_trace} can be used.
\begin{small}
\begin{verbatim}
>>> axes = sbijax.plot_posterior(inference_results: InferenceData)
>>> axes = sbijax.plot_trace(inference_results: InferenceData)
\end{verbatim}
\end{small}

\begin{figure}[h]
    \centering
    \includegraphics[width=1\textwidth]{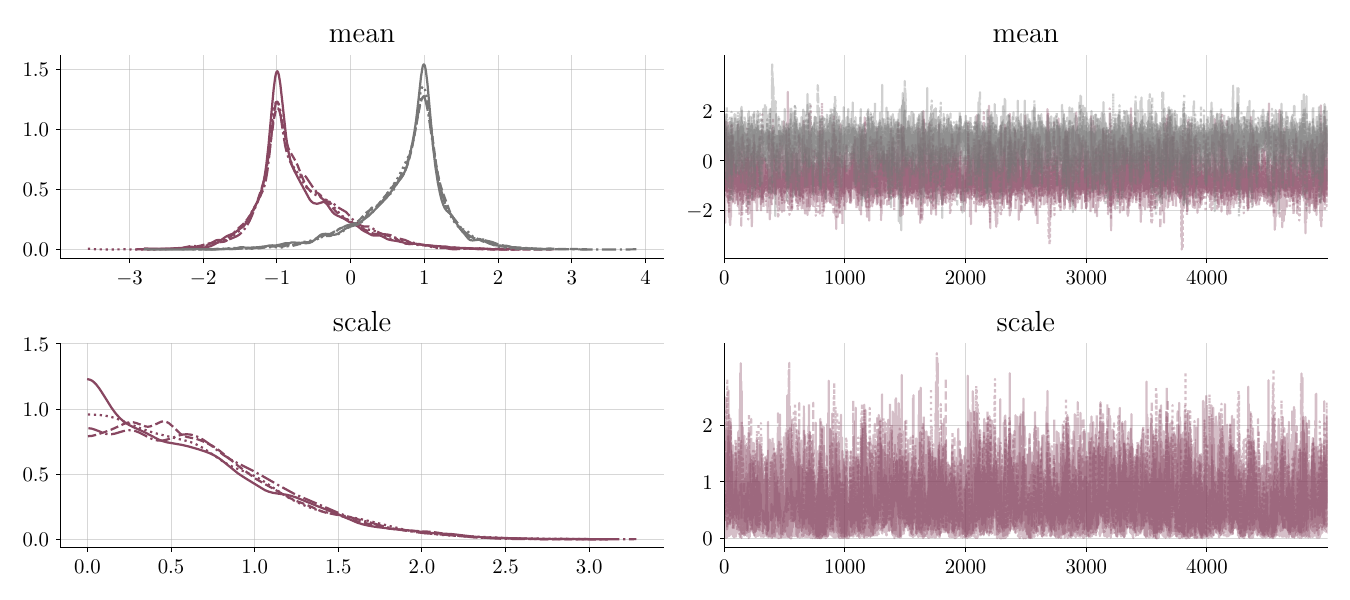}
    \caption{Marginal posterior density and trace plots. Each variable is visualized separately. Multivariate parameter like the mean are shown in different colors in a panel. The titles of the figures respect the variable names of the generative model (see Section~\ref{sec:sbijax-model_definition}).}
    \label{fig:trace-plot}
\end{figure}

The functions only take a single argument \texttt{inference\_results} and returns a list of \texttt{axis} objects (see Figure~\ref{fig:trace-plot} for an example).

\begin{figure}
\begin{subfigure}[b]{1\textwidth}
\centering
   \includegraphics[width=.8\linewidth]{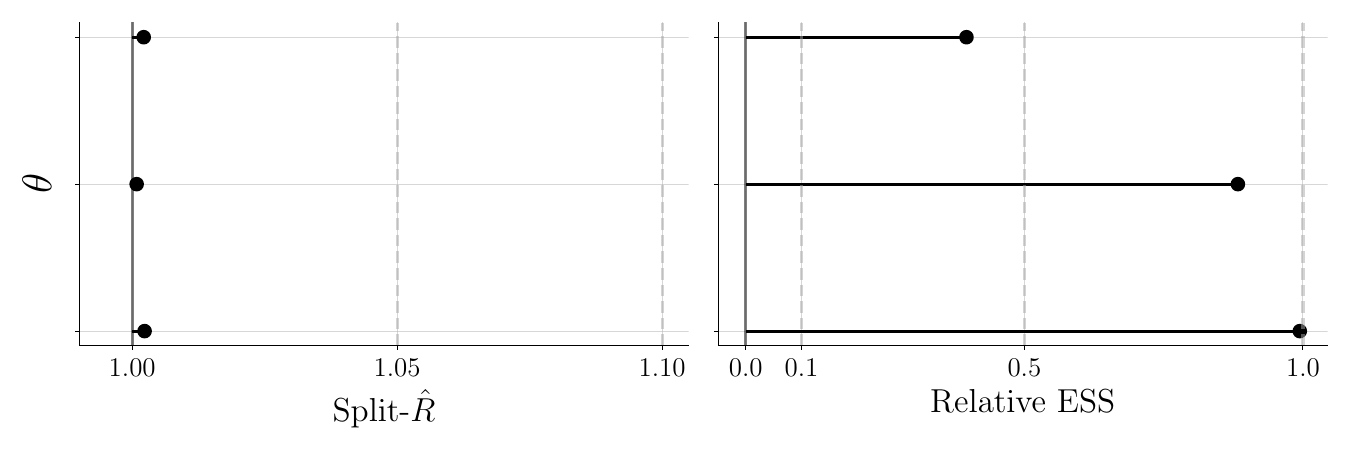}
   \caption{Split-$\hat{R}$ and relative effective sample size. The two statistics are shown  concisely for all parameters jointly. Following the convention in \texttt{bayesplot}, we introduce thresholds for split-$\hat{R}$ and relative ESS to help the user identify good mixing. The values are arbitrary and not theoretically grounded.}
   \label{fig:rhat-and-ress}
\end{subfigure}
\begin{subfigure}[b]{1\textwidth}        
   \centering
   \includegraphics[width=\textwidth]{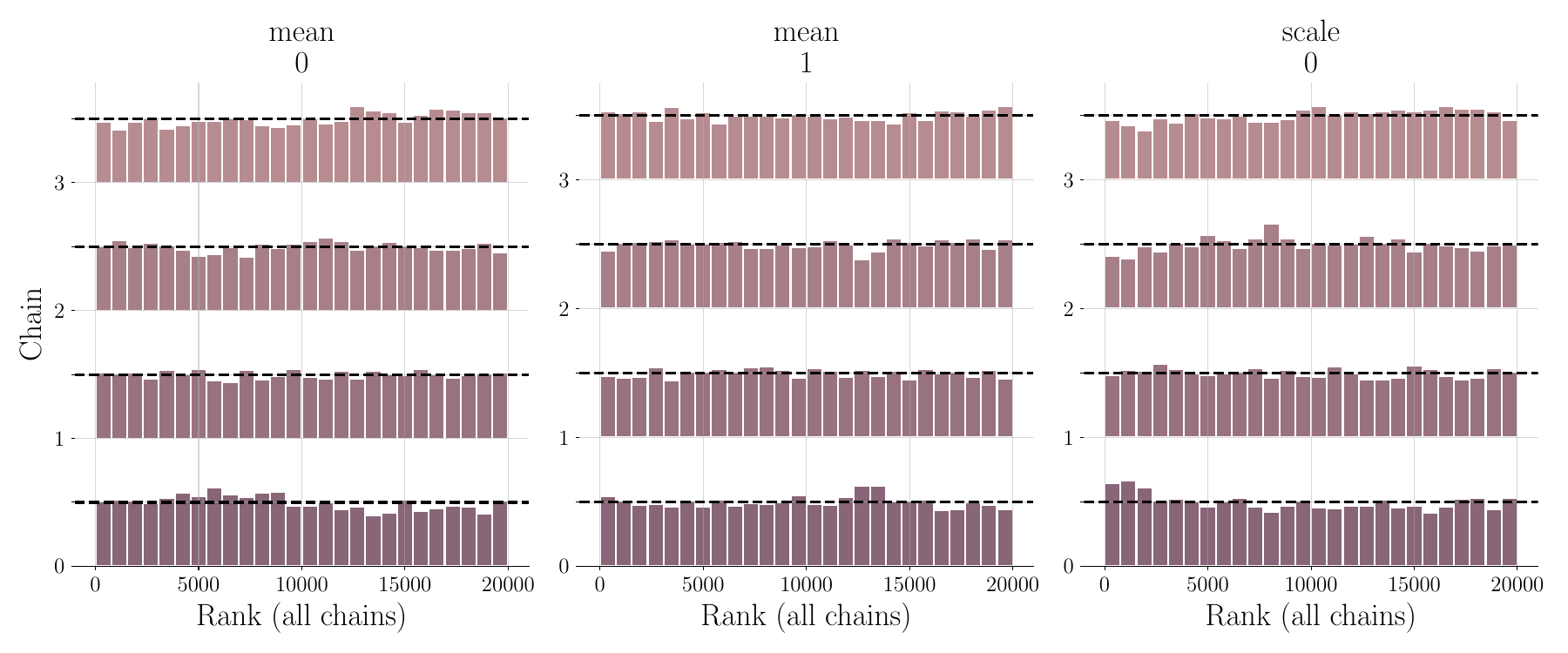}
   \caption{Rank statistics plot. Rank plots show the ranked posterior draws for each chain aand parameter separately. If all chains target the same posterior, ranks are supposed to be uniform. If rank plots look similar for a parameter, good mixing of chains can be assumed.}
   \label{fig:ranks}
\end{subfigure}
\begin{subfigure}[b]{1\textwidth}
    \centering
   \includegraphics[width=.95\linewidth]{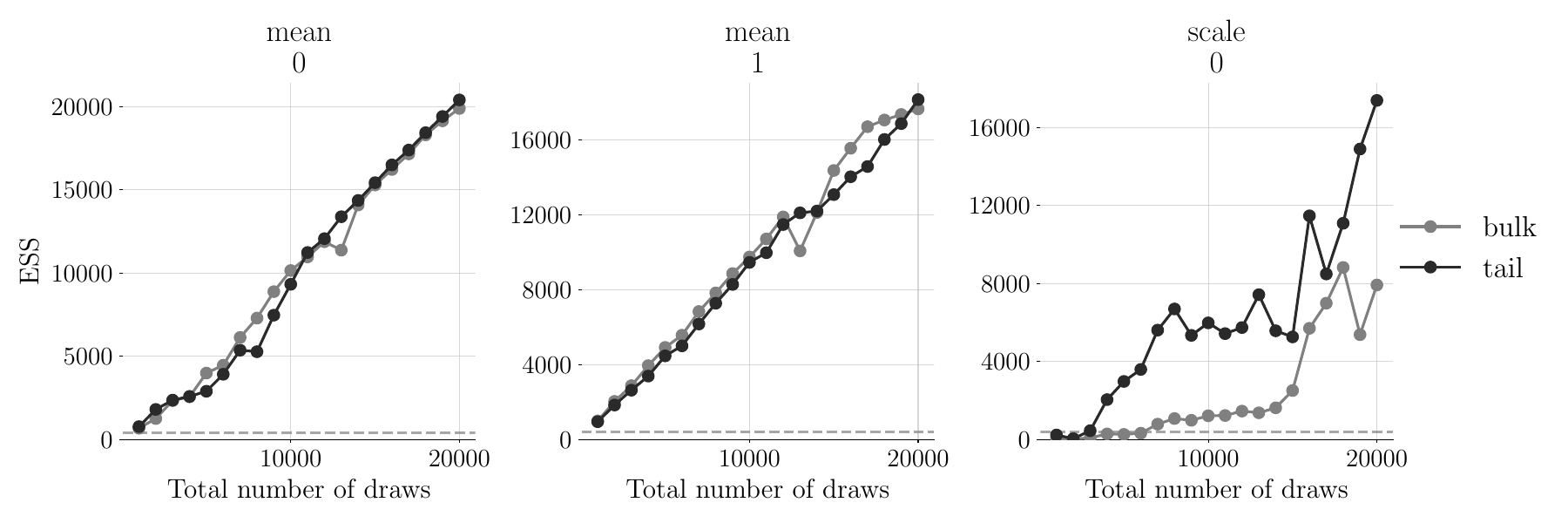}
   \caption{Effective sample size plot. The figures show the evolution of the effective sample size for each parameter during MCMC sampling. Different colors represent the bulk ESS and the tail ESS which is defined as the minimum of the effective sample sizes for $5\%$ and $95\%$ quantiles.}
   \label{fig:ess}
\end{subfigure}
\caption{Split-$\hat{R}$, rank and effective sample size plots.}
\label{fig:ranks-and-ess}
\end{figure}

For visualization of MCMC model diagnostics, specifically the split-$\hat{R}$ and the relative ESS, the functions \texttt{plot\_rhat\_and\_ress}, \texttt{plot\_rank}, or \texttt{plot\_ess} can be used:
\begin{small}
\begin{verbatim}
>>> axes = sbijax.plot_rhat_and_ress(inference_results: InferenceData)
>>> axes = sbijax.plot_rank(inference_results: InferenceData)
>>> axes = sbijax.plot_ess(inference_results: InferenceData)
\end{verbatim}
\end{small}
As before, the sole argument of the functions is an \texttt{InferenceData} object. The function \texttt{plot\_rhat\_and\_ress} is inspired by the visualizations of the {R} package \texttt{bayesplot} \citep{gabry2019visualization,gabry2024bayesplot} and gives a minimal overview of sampler diagnostics. It shows the split-$\hat{R}$ and the relative ESS for all variables $\theta$ on the pooled samples of all MCMC chains (Figure~\ref{fig:rhat-and-ress}). The function \texttt{plot\_rank} visualizes the rank statistics for each chain separately (Figure~\ref{fig:ranks}; we refer the reader to \citet{vehtari2021split} for details), while \texttt{plot\_ess} shows the chain-averaged evolution of the bulk and tail effective sample sizes against the number of MCMC iterations (Figure~\ref{fig:ess}).

\subsection{Sequential inference}
\label{sec:sequential-training}

The neural SBI algorithms implemented in \texttt{sbijax} allow to sequentially infer the posterior distribution for a fixed measurement $y_{\text{obs}}$. In that case, amortized inference, i.e., inference for any observable given a trained surrogate model, is not possible any more. Empirically, it has been shown that this increases the accuracy of the approximation $\hat{\pi}(\theta | y_{\text{obs}})$  (see Section~\ref{sec:background-sequential-inference}). In \texttt{sbijax}, sequential training is possible by iteratively sampling data and then re-fitting the surrogate model, for instance:
\begin{small}
\begin{verbatim}
>>> from sbijax.util import stack_data
>>> n_rounds = 5
>>> data, params = None, {}
>>> for i in range(n_rounds):
...     new_data, _ = model.simulate_data(
...         jr.fold_in(jr.PRNGKey(1), i),
...         params=params,
...         observable=y_obs,
...         data=data,
...     )
...     data = stack_data(data, new_data)
...     params, info = model.fit(jr.fold_in(jr.PRNGKey(2), i), data=data)
>>>
>>> inference_results, diagnostics = model.sample_posterior(
...    jr.PRNGKey(3), params, observable
... )    
\end{verbatim}
\end{small}
In the code excerpt above, we used the function \texttt{model.simulate\_data} with two previously unused additional arguments (cf. Section~\ref{sec:sbijax-training_and_inference}): the observation $y_{\text{obs}}$ (called \texttt{observable}) and a parameter argument (called \texttt{params}). These can be provided additionally if the parameter values are sampled from the surrogate posterior, i.e., instead of sampling
\begin{equation*}
y_n \sim \pi(y|\theta_n), \quad \theta_n \sim \pi(\theta)
\end{equation*}
we sample
\begin{equation*}
y_n \sim \pi(y|\theta_n), \quad \theta_n \sim \hat{\pi}(\theta | y_{\text{obs}})
\end{equation*}
The newly sampled data is then appended to the existing data set using the function \texttt{stack\_data} before the entire data set is used
to train the surrogate model. After $R$ rounds, the procedure terminates and samples from the approximate posterior can be taken.

In \texttt{sbijax}, each method supports sequential inference. However, algorithms that have not been introduced with sequential training (in their original publications), emit a warning to the user that care must be taken.

\subsection{Automatic computation of summary statistics}

Computational inference of posterior distributions in high-dimensional data spaces can be complicated for both SMC-ABC methods, since good summary statistics have to be found but do not necessarily exist, and neural SBI methods, since, e.g., a density for the likelihood function has to estimated. The \texttt{sbijax} packages offers functionality to automatically compute summaries with neural networks that map data to from their high-dimensional representation to a lower-dimensional embedding \citep{chen2021neural,chen2023learning}. The trained summary network can be used in a down-stream analysis to learn models on the summarized data, e.g., using SMC-ABC or neural SBI methods. For instance, a workflow that learns summary statistics automatically using {NASS} \citep{chen2021neural} and then constructs a function to summarize data for, e.g., SMC-ABC, is shown below:
\begin{small}
\begin{verbatim}
>>> n_dim_summary, hidden_sizes = 1, (64, 64)
>>> neural_network = make_nass_net(n_dim_summary, hidden_sizes)
>>>
>>> fns = prior_fn, simulator_fn
>>> summ_model = NASS(fns, neural_network)
>>> 
>>> data = model.simulate_data(seed=jr.PRNGKey(1), n_simulations=10_000)
>>> params, loss = summ_model.fit(jr.PRNGKey(2), data=data)
>>>
>>> def summary_fn(y):
...     return summ_model.summarize(y, params)
\end{verbatim}
\end{small}

The constructor of the summary method {NASS}, and the related method {NASSS} (see table~\ref{tab:all-methods}), takes as a first argument a tuple of prior and simulator functions and as a second argument a summary neural network. The network can either be constructed using the function \texttt{make\_nass\_net} or using a custom user-defined function. In both cases it needs to reduce the data to a desired dimensionality which in the example above is one. The {NASS} algorithm is then trained using simulated data $\{(y_n, \theta_n) \}_{i=1}^N$ from the generative model (cf. Section~\ref{sec:sbijax-training_and_inference}). The training method \texttt{fit} returns a set of neural network weights, above called \texttt{params}, and training and validation set loss profiles (cf. Section~\ref{sec:sbijax-training_and_inference}). 

\subsection{Neural networks}

The declaration of SBI objects in \texttt{sbijax} requires to first define a neural network architecture that is used for inference of approximate posterior distributions, likelihood functions, or likelihood-to-evidence ratios. 

The high-level interface of \texttt{sbijax} provides user-friendly functions to define these architectures (table~\ref{tabl:neural-networks}). For instance, methods that estimate probability density functions, such as NLE, SNLE and NPE, require either normalizing flows or mixture density networks as neural network models. In that case, the functions \texttt{make\_mdn}, \texttt{make\_maf} or \texttt{make\_spf} can be used. For methods that estimate likelihood-to-evidence rations, such as NRE, classifier networks, like MLPs (\texttt{make\_mlp}) and ResNets (\texttt{make\_resnet}; \citet{he2016deep}) are required. Methods, that estimate vector fields, such as FMPE and CMPE can use the functions \texttt{make\_cnf} and \texttt{make\_cm}, respectively, to build neural networks. 

\begin{table}
\centering
\caption{{Methods for constructing neural networks.} The high-level API of \texttt{sbijax} provides several functions for constructing neural networks. Depending on the SBI algorithm to be applied, different neural networks are admissible.}
\begin{tabular}{p{40mm}llp{40mm}}
\toprule
&\textbf{Method name} & \textbf{Algorithms} & \textbf{Main reference} \\
\midrule
Mixture density net    & \texttt{make\_mdn}  & \texttt{(S)NLE, NPE}  & \citet{bishop1994mixture} \\
Affine MAF    & \texttt{make\_maf}   & \texttt{(S)NLE, NPE}  & \citet{papamakarios2017masked} \\
Spline coupling flow                 & \texttt{make\_spf}& \texttt{(S)NLE, NPE} & \citet{durkan2019neural,dinh2017density}\\
Continuous flow                      & \texttt{make\_cnf} & \texttt{FMPE} & \citet{chen2018neurao}\\
MLP                                  & \texttt{make\_mlp} & \texttt{NRE}  & -- \\
ResNet                               & \texttt{make\_resnet}& \texttt{NRE} & \citet{he2016deep}\\
Consistency model                    & \texttt{make\_cm}& \texttt{CMPE} & \citet{song2023consistency}\\
NASS                                 & \texttt{make\_nass\_net}& \texttt{NASS} & \citet{chen2021neural}\\
NASSS                                & \texttt{make\_nasss\_net}& \texttt{NASSS} & \citet{chen2023learning}\\
\bottomrule
\end{tabular}
\label{tabl:neural-networks}
\end{table}

\subsection{Sampling algorithms}

SBI algorithms that do not directly approximate the posterior distribution, e.g., NLE, NRE or NASS, need to sample from the surrogate posterior distribution using MCMC methodology. We provide several common samplers building on the high-quality implementations of the Python package \texttt{BlackJAX} \citep{cabezas2024blackjax}. Table~\ref{tabl:samplers} lists the samplers that are currently implemented. Per default, \texttt{sbijax} utilizes a No-U-Turn sampler (NUTS; \citet{hoffman2014no}) for all algorithms that utilize MCMC sampling.

\begin{table}
\centering
\caption{{MCMC algorithm for sampling from the approximate posterior distribution.} \texttt{sbijax} makes use of the high-performance implementations provided by the \texttt{BlackJAX} sampling library.}
\begin{tabular}{ll}
\toprule
\textbf{MCMC sampler} & \textbf{Main reference} \\
\midrule
Rosenbluth-Metropolis-Hastings sampler  & \citet{brooks2011mcmc}            \\
Metropolis-adjusted Langevin algorithm   & \citet{brooks2011mcmc}        \\   
Slice sampler & \citet{neal2003slice}  \\
No-U-Turn sampler  &  \citet{hoffman2014no} \\
\bottomrule
\end{tabular}
\label{tabl:samplers}
\end{table}

\subsection{Continuous integration and package installation}

We employ conventional software engineering protocols to ensure high-quality, safe, and correct code. Specifically, we make use of GitHub workflows to automatically run unit tests, lint source code, notify about potential security issues and bugs, compute code coverage, and compute code quality. We use \href{www.readthedocs.com}{\texttt{readthedocs.com}} to automatically host code documentation.

The package \texttt{sbijax} is available from the {Python} package repository PyPI and can be installed from the command line using:
\begin{small}
\begin{verbatim}
pip install sbijax
\end{verbatim}
\end{small}

\section{Examples}
\label{sec:examples}
In the following, we give an overview of several implemented algorithms from Table~\ref{tab:all-methods} to illustrate \texttt{sbijax}. The documentation of all algorithms with detailed usage instructions can be found online at \href{https://sbijax.rtfd.io}{\texttt{sbijax.rtfd.io}}. Additional examples can be found in Appendix~\ref{app:additional-algorithm-examples}.

We motive \texttt {sbijax} using a common experimental benchmark model from the SBI literature that has a complex multi-modal posterior distribution but a simple likelihood function (commonly referred to as SLCP model). The generative model has the following form:
\begin{equation}
\begin{split}
\theta_i &\sim \text{Uniform}(-3, 3) \; \text{for} \; i=1, \dots, 5\\
\mu( {\theta}) &= (\theta_1, \theta_2), \phi_1 = \theta_3^2 , \phi_2 = \theta_4^2 \\
\Sigma( {\theta} ) &=
\begin{pmatrix}
\phi_1^2 & \text{tanh}(\theta_5) \phi_1 \phi_2 \\
\text{tanh}(\theta_5) \phi_1 \phi_2 & \phi_2^2
\end{pmatrix}\\
{y}_j | {\theta}  &\sim \mathcal{N}({y}_j; \mu( {\theta}), \Sigma( {\theta})) \; \text{for} \; j=1, \dots, 4\\
{y} &= [{y}_1, \dots, {y}_4]^T
\end{split}
\end{equation}
The parameter of interest $\theta \in \mathbb{R}^5$ parameterizes the mean and covariance matrix of a bivariate Gaussian. Sampling from the observation model yields an eight-dimensional random variable $y \in \mathbb{R}^8$ (we refer to Appendix~\ref{app:additional-material-for-slcp} for the transcription of the generative model in {Python} code). 

We are interested in inferring the posterior distribution $\pi(\theta|y_{\text{obs}})$, and evaluate three different SBI algorithms, the dimensionality-reducing surjective NLE (SNLE),
SMC-ABC using neural sufficient statistics, and FMPE. 
As an observation $y_{\text{obs}}$ we use the same as in \citet{papamakarios2019sequential} (see Appendix~\ref{app:additional-material-for-slcp} for details). For comparison, we also infer the posterior distribution with a slice sampler, since the generative model, as in the example in Section~\ref{sec:package}, admits a likelihood function that is tractable to compute.  Before discussing the results, we show below how to use the computational methods.

\subsection{Surjective neural likelihood estimation}
\label{sec:examples-nle}
To do inference using neural likelihood estimation, the user first needs to specify a conditional density model which will be used to estimate a conditional probability density function $q(y, \theta)$ for the data.
\begin{small}
\begin{verbatim}
>>> n_dim_data = 8
>>> n_layer_dimensions, hidden_sizes = (8, 8, 5, 5, 5), (64, 64)
>>> neural_network = make_maf(
...     n_dim_data, 
...     n_layer_dimensions=n_layer_dimensions,
...     hidden_sizes=hidden_sizes
... )
>>>
>>> fns = prior_fn, simulator_fn
>>> nle = SNLE(fns, neural_network)
\end{verbatim}
\end{small}
In the code excerpt above, a surjective affine masked autoregressive flows, consisting of five flow layers, is constructed to model an eight-dimensional space and reduce it to a five-dimensional space at the third layer. The \texttt{make\_maf} function of \texttt{sbijax} automatically recognizes which type of flow is supposed to be built when a tuple that specifies the layer dimensionalities is provided (instead of an integer that specified the number of layers (c.f. Section~\ref{sec:sbijax-alg_definition}).

We then do sequential inference in $R=15$ rounds. The {Python} code describing the inferential problem is shown below.
\begin{small}
\begin{verbatim}
>>> data, snle_params = None, {}
>>> for i in range(15):
...     data, _ = snle.simulate_data_and_possibly_append(
...         jr.fold_in(jr.PRNGKey(1), i),
...         params=snle_params,
...         observable=y_obs,
...         data=data
...     )
...     snle_params, info = snle.fit(
...         jr.fold_in(jr.PRNGKey(2), i), data=data
...     )
>>>
>>> snle_inference_results, diagnostics = snle.sample_posterior(
...     jr.PRNGKey(3), snle_params, y_obs, n_samples=5000, n_warmup=2500, n_chains=10
... )
\end{verbatim}
\end{small}

Before the training loop, we create two objects, called \texttt{data} and \texttt{snle\_params}. In each round, the approximate posterior of the previous round $\hat{\pi}^{r-1}(\theta|y_{\text{obs}})$ is used to draw a parameter sample and appended to the object \texttt{data}. Then, the surrogate model for the likelihood is fit on the entire data set and the parameters of the neural network which are returned are saved in \texttt{snle\_params}. Finally, upon conclusion of all training rounds, a posterior sample is drawn using, by default, a No-U-Turn sampler on the surrogate posterior. Here, we sample from ten separate chains in parallel.

\subsection{Flow matching posterior estimation}
Posterior inference using neural posterior estimation approaches requires defining a pushforward distribution on the parameter-space. In the case of flow matching posterior estimation, \texttt{FMPE}, which we illustrate below, a continuous normalizing flow has to be constructed. This can be done using the high-level interface of \texttt{sbijax}. Since continuous NFs can use any neural network $h: \mathbb{R}^{d_y} \times \mathbb{R}^{d_\theta} \rightarrow \mathbb{R}^{d_\theta}$ and consequently do not suffer from the same constraints as conventional NFs, we modify the architecture slightly relative to the NLE example (Section~\ref{sec:examples-nle}).
\begin{small}
\begin{verbatim}
>>> n_dim_theta = 5
>>> n_layers, hidden_size = 5, 128
>>>
>>> neural_network = make_cnf(n_dim_theta, n_layers, hidden_size)
>>> fns = prior_fn, simulator_fn
>>> fmpe = FMPE(fns, neural_network)
\end{verbatim}
\end{small}
Posterior samples can be drawn by first simulating data, fitting the model, and then sampling from the pushforward. Since in the original publication, \citet{wildberger2023flow} did not propose a sequential training procedure for FMPE, in the following code example we will be using amortized inference.

\begin{small}
\begin{verbatim}
>>> data, _ = fmpe.simulate_data(
...     jr.PRNGKey(1),
...     n_simulations=20_000,
... )
>>> fmpe_params, info = fmpe.fit(
...     jr.PRNGKey(2),
...     data=data
... )
>>> fmpe_inference_results, diagnostics = fmpe.sample_posterior(
...     jr.PRNGKey(3), fmpe_params, y_obs, n_samples=25_000
... )
\end{verbatim}
\end{small}

\subsection{Sequential Monte Carlo ABC using neural sufficient statistics}

SMC-ABC requires the definition of functions to compute summary statistics for a data point, and the distance between the summaries of the observed and simulated data. For the eight-dimensional SLCP example above, we will use \texttt{NASS} to automatically compute summary statistics \citep{chen2021neural}. The \texttt{NASS} constructor requires a tuple of prior and simulator functions and a trainable summary network. We can use the \texttt{make\_nass\_net} function to automatically construct such a network where the arguments specify the architecture of the network, and then train the neural network with the same function calls as before.
\begin{small}
\begin{verbatim}
>>> n_embedding_dim, hidden_sizes = 5, (64, 64)
>>> neural_network = make_nass_net(n_embedding_dim, hidden_sizes)
>>>
>>> fns = prior_fn, simulator_fn
>>> model_nass = NASS(fns, neural_network)
>>>
>>> data, _ = model_nass.simulate_data(jr.PRNGKey(1), n_simulations=20_000)
>>> params_nass, _ = model_nass.fit(jr.PRNGKey(2), data=data)
\end{verbatim}
\end{small}
With a trained summary network, below we define a function, called \texttt{summary\_fn}, which uses the summary network to compute statistics of the data. The Euclidean distance is computed between the summary statistics of the observed and simulated data sets.
\begin{small}
\begin{verbatim}
>>> def summary_fn(y):
...     s = model_nass.summarize(params_nass, y)
...     return s
>>>
>>> def distance_fn(y_simulated, y_observed):
...     diff = y_simulated - y_observed
...     dist = jax.vmap(lambda el: jnp.linalg.norm(el))(diff)
...     return dist
\end{verbatim}
\end{small}

SMC-ABC infers the posterior distribution sequentially by reducing the allowed distance in each round which ultimately encourages the drawn particles to converge to the true posterior distribution. In \texttt{sbijax}, the SMC-ABC posterior is drawn as shown below.
\begin{small}
\begin{verbatim}
>>> fns = prior_fn, simulator_fn
>>> smc = SMCABC(fns, summary_fn, distance_fn)
>>>
>>> smc_inference_results, _ = smc.sample_posterior(
...     jr.PRNGKey(3), y_obs, n_rounds=10
... )
\end{verbatim}
\end{small}
As for all other algorithms, the \texttt{SMCABC} constructor takes, as a first argument, the tuple of prior and simulator functions. The second argument is the summary function, \texttt{summary\_fn}, and the third one the distance function, \texttt{distance\_fn}. 

To infer posterior samples, the function \texttt{sample\_posterior} is used with the argument \texttt{n\_rounds} used to specify the number of sequential rounds.

\begin{figure}
  \centering
        \begin{subfigure}[b]{0.49\textwidth}
            \centering
            \includegraphics[width=\textwidth]{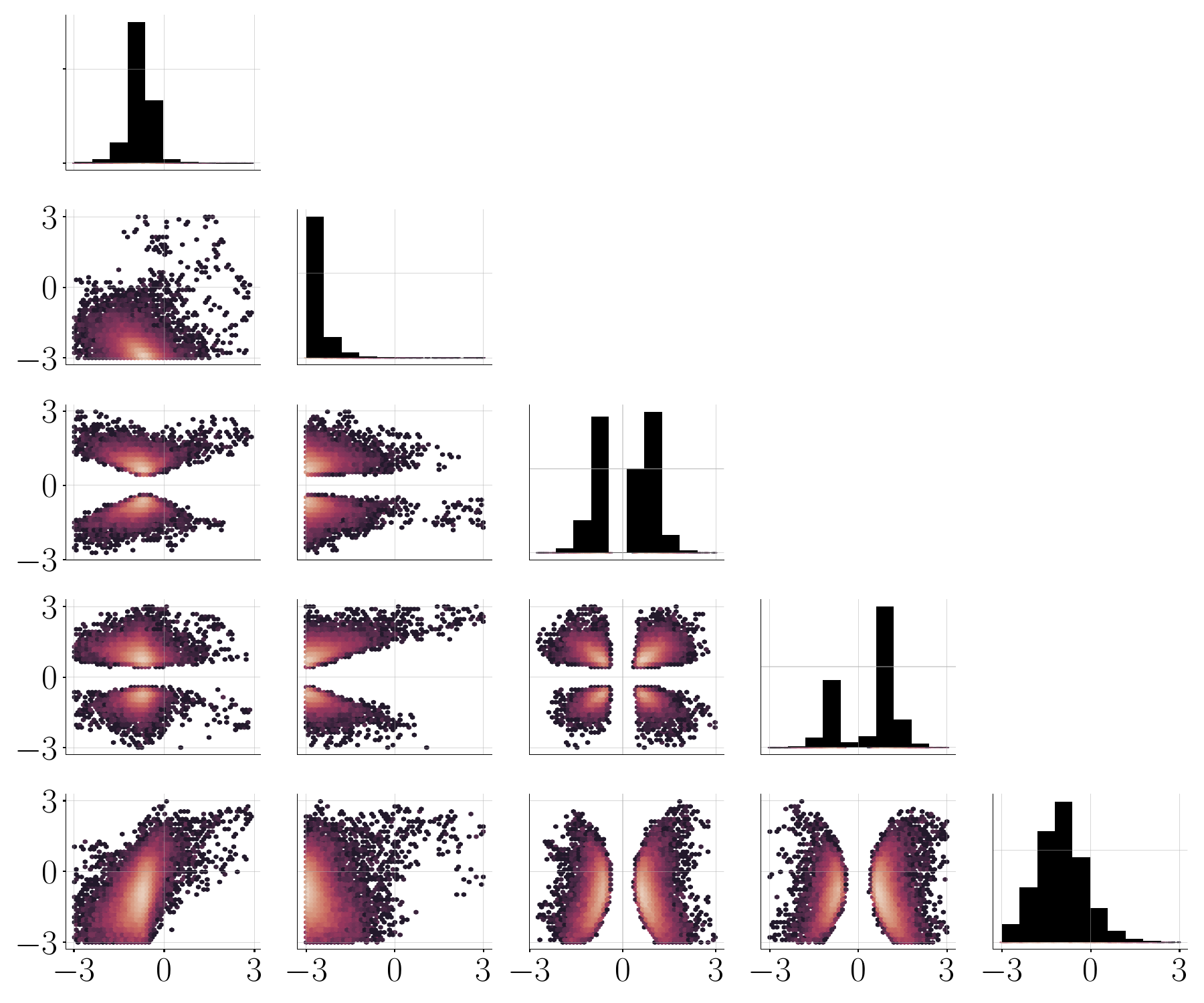}
            \caption{{Slice sampling posterior.}}
            \label{fig:posterior_pairs-slice}
        \end{subfigure}
        \hfill
        \begin{subfigure}[b]{0.49\textwidth}   
            \centering 
            \includegraphics[width=\textwidth]{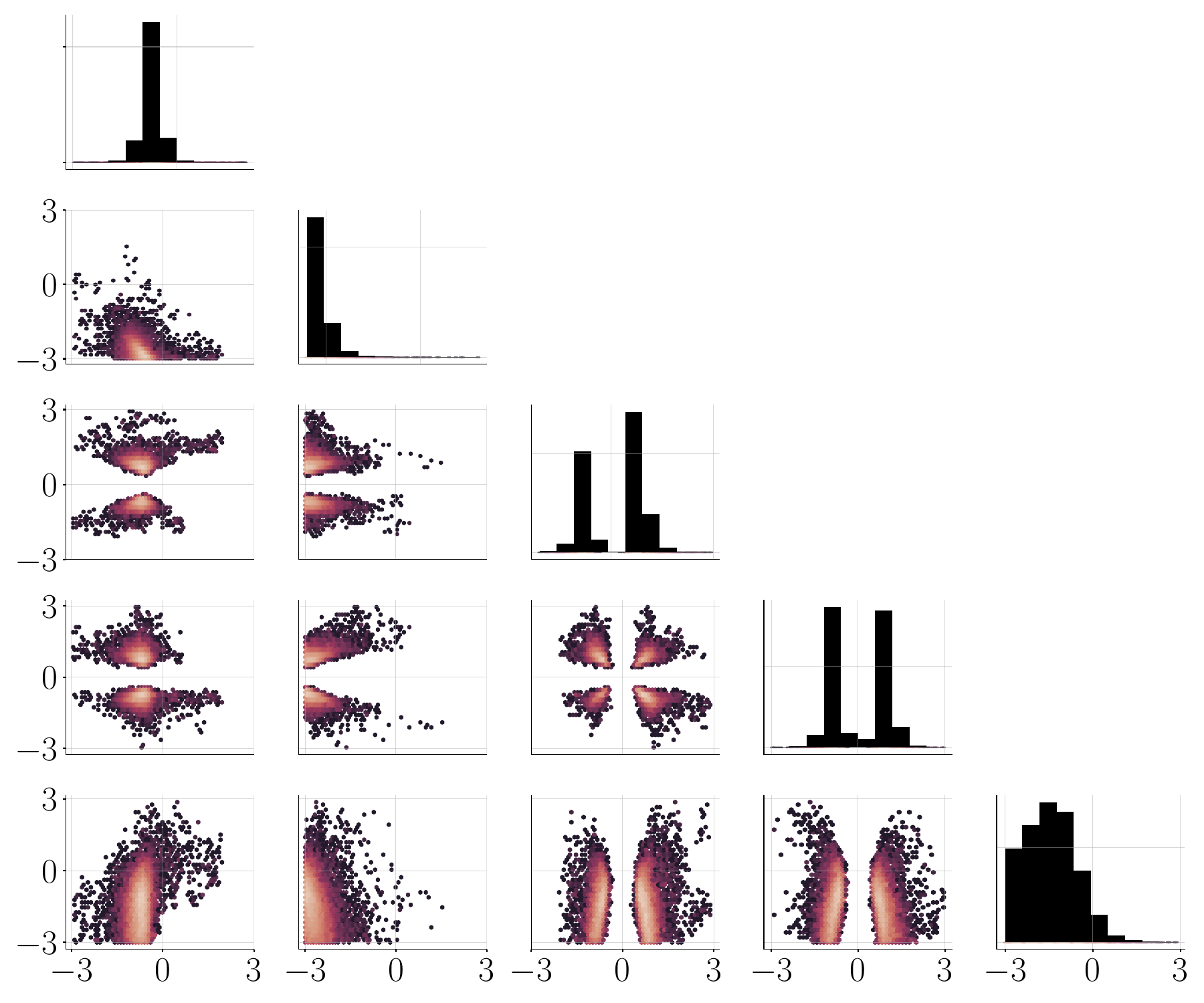}
            \caption{{SNLE posterior.}}
            \label{fig:posterior_pairs-nle}
        \end{subfigure}        
        \vskip\baselineskip         
        \begin{subfigure}[b]{0.49\textwidth}   
            \centering 
            \includegraphics[width=\textwidth]{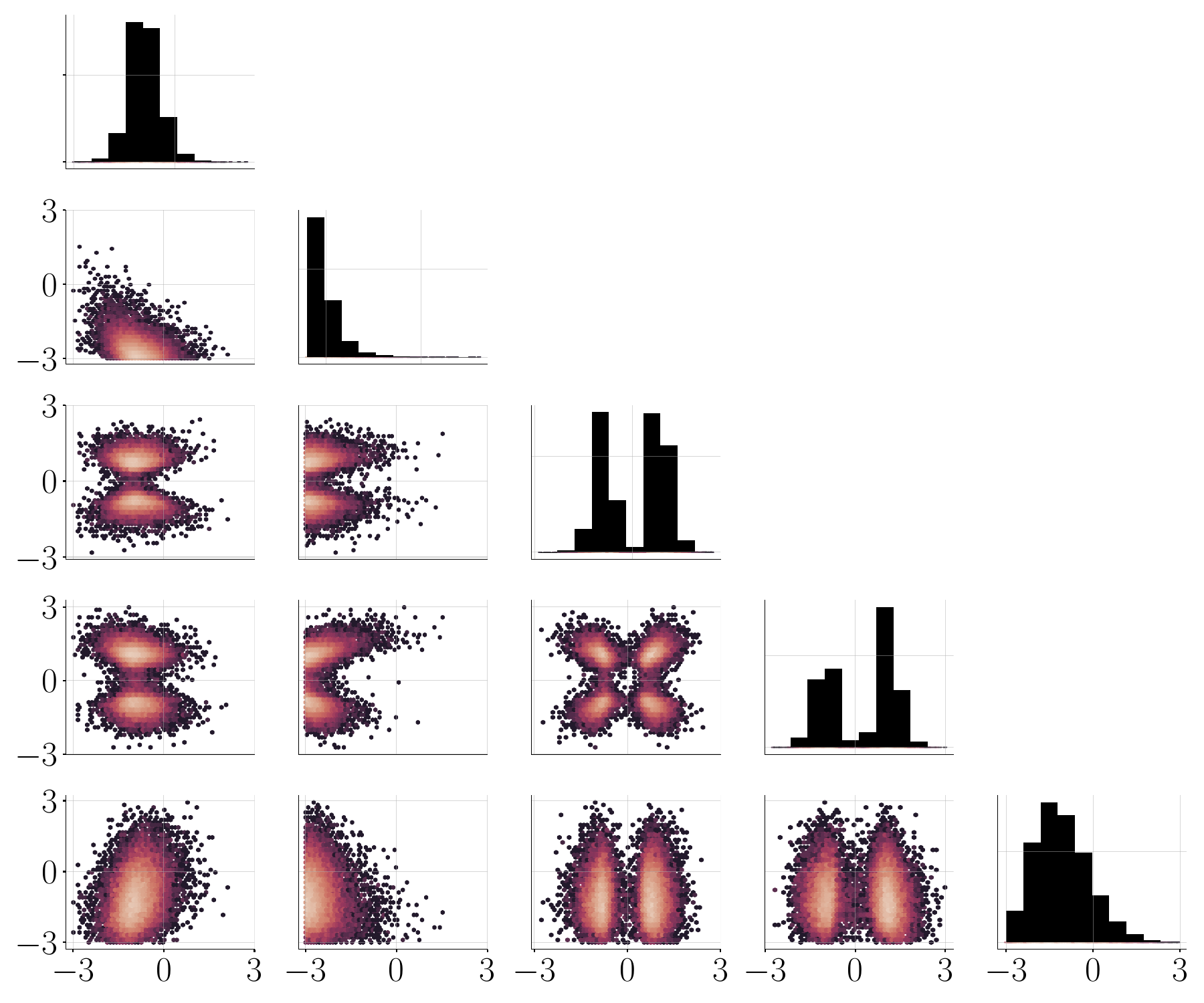}
            \caption{{FMPE posterior.}}
            \label{fig:posterior_pairs-fmpe}
        \end{subfigure}
        \hfill
        \begin{subfigure}[b]{0.49\textwidth}  
            \centering 
            \includegraphics[width=\textwidth]{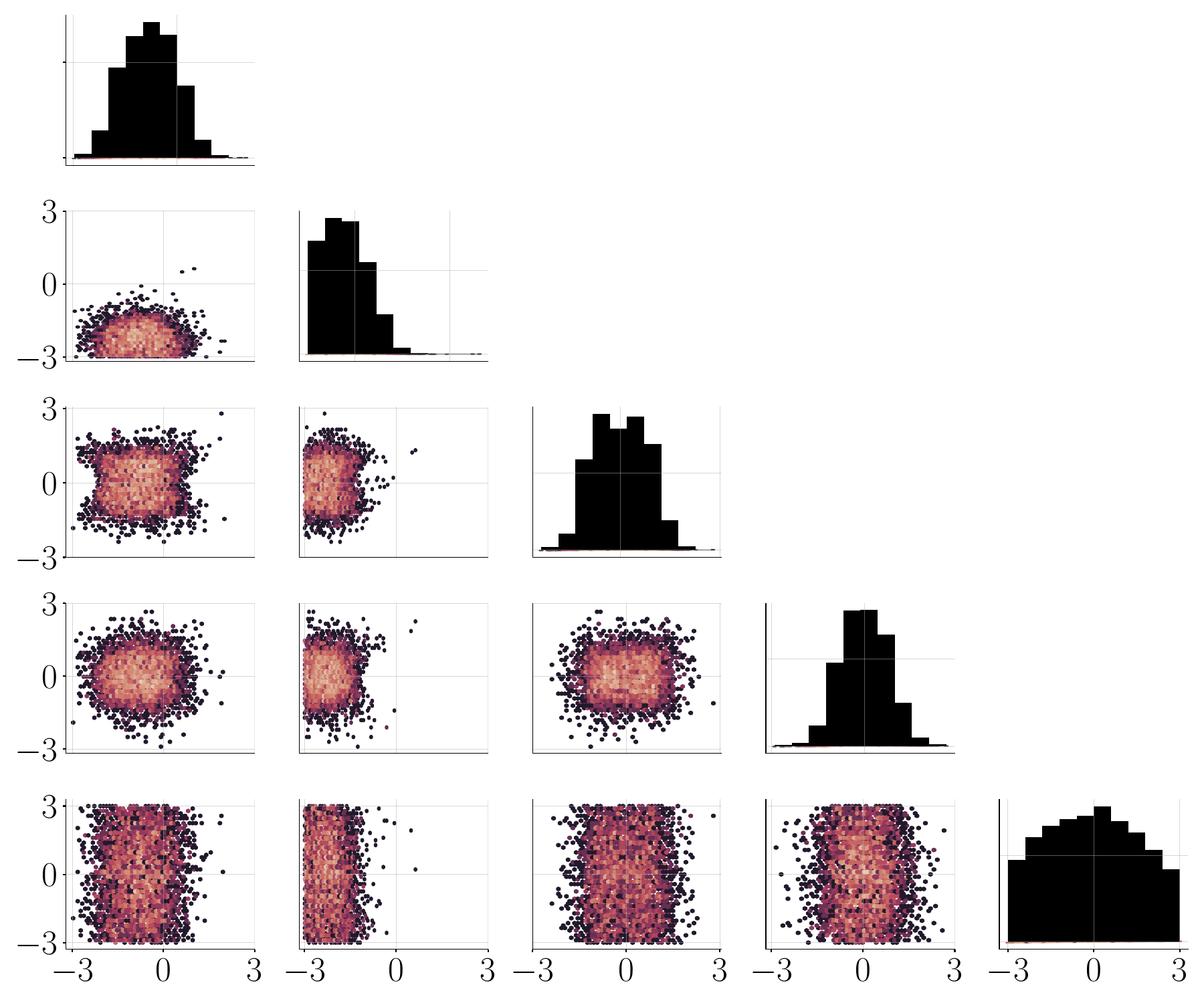}
            \caption{{SMC-ABC+NASS posterior.}}
            \label{fig:posterior_pairs-smc}
        \end{subfigure}
        \caption{Posterior pair plots and marginal distributions. For this benchmark model (SLCP), SNLE achieves the best approximation to the true posterior (when comparing to posterior distribution inferred using the slice sampler). SMC-ABC+NASS and FMPE show worse performance.}
\label{fig:posterior_pairs}
\end{figure}
\subsection{Results}
We compare the three approximate approaches to the posterior inferred using a slice sampler \citep{neal2003slice}. We draw samples from the true posterior using $10$ separate chains of length $10\ 000$ of which we discard the first $5\ 000$ as warmup samples (Figure~\ref{fig:posterior_pairs-slice}; MCMC model diagnostics can be found in Figures~\ref{fig:slcp-slice-rhat_ress} and \ref{fig:slcp-slice-tre-plot} in Appendix~\ref{app:additional-material-for-slcp}). 

The MCMC sample demonstrates the complexity of approximating the posterior model. The bi-variate marginals of some elements of $\theta$ are multi-modal which generally poses significant challenges even for MCMC samplers. In this experiment, the density model of NLE approximates the true posterior distribution the best (Figure~\ref{fig:posterior_pairs-nle}). This is expected since a sequential sampling scheme was chosen. The posterior samples of SMC-ABC+NASS and FMPE are graphically farther away from the true posterior. SMC-ABC+NASS, with the selected parameter setup and computational budget, does not manage to reveal the multi-modality of the posterior (Figure~\ref{fig:posterior_pairs-smc}), while FMPE comes very close even with an amortized inferential scheme (Figure~\ref{fig:posterior_pairs-fmpe}).

\section{Experimental data use case}
\label{sec:real-data}
Finally, we apply \texttt{sbijax} to an experimental data set of electroencephalogram (EEG) recordings. The data consists of time series measurements from multiple test subjects who are asked to keep their eyes opened or closed while the EEG is conducted. When exempt of visual stimuli, subjects are expected to show a prominent occipital dominant rhythm (commonly called occipital alpha wave). With open eyes these waves are expected to be reduced. The time series are measured over eight seconds at a frequency of $512Hz$. We resample the trajectory at $128Hz$ which yields 1025 time points per time series (see \citet{cattan2018} for details).

\begin{figure}[h!t]
  \centering
        \begin{subfigure}[b]{0.475\textwidth}
            \centering
            \includegraphics[width=\textwidth]{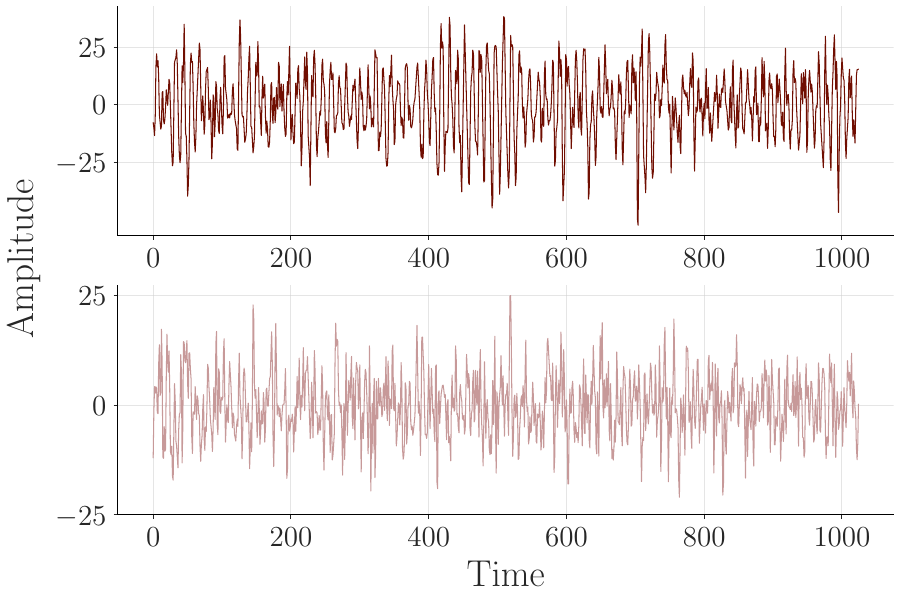}
            \caption{Raw EEG recordings.}
            \label{fig:eeg-raw}
        \end{subfigure}vr
        \hfill
        \begin{subfigure}[b]{0.475\textwidth}   
            \centering 
            \includegraphics[width=\textwidth]{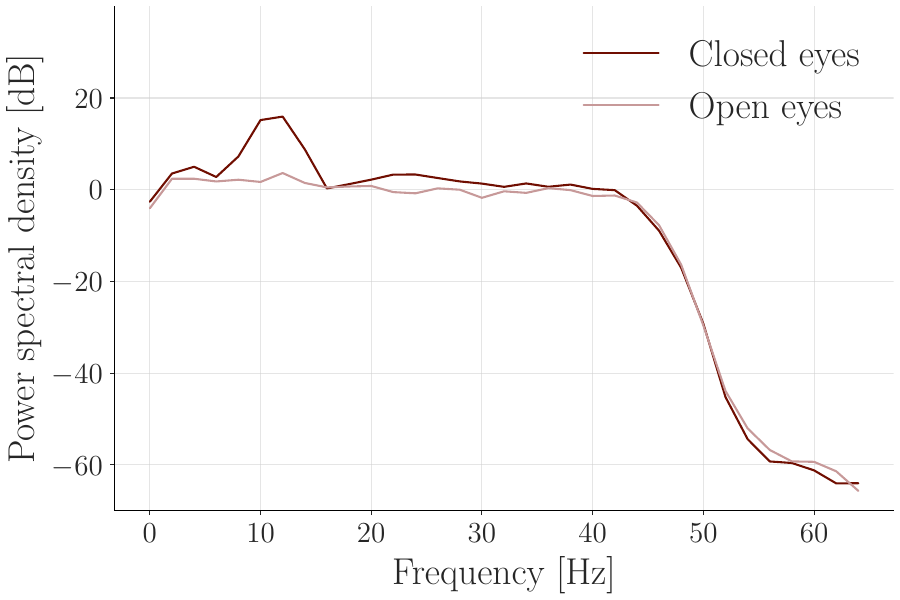}
            \caption{EEG recordings periodogram.}
            \label{fig:eeg-periodogram}
        \end{subfigure}        
        \caption{EEG measurements from a subject with eyes closed and open. (a) The raw trajectories from the EEG consist of $1025$ time points that are measured at $512Hz$ and resampled to $128Hz$. (b) The power spectral density evaluated at $33$ frequency bins of both trajectories in decibels.}
\label{fig:eeg-data}
\end{figure}

Following \citet{rodrigues2021hnpe}, we select a random subject and a pair of measurements from their EEG recordings, one with open eyes and one with closed eyes, and aim to infer the posterior distributions of biologically interesting parameters for each of the two measurements (see Figure~\ref{fig:eeg-raw} for the two selected trajectories and Figure~\ref{fig:eeg-raw-full} for all trajectories). Below, we will first describe a suitable simulator function and a prior model for its parameters, and then infer the posterior distributions for the two measurements using an amortized neural estimator.

\subsection{Prior and simulator functions}
We utilize a stochastic version of the Jansen-Rit neural mass model (JRNMM; \citet{ableidinger2017stochastic}) as a simulator function to describe the data generating mechanism of an EEG. The stochastic JRNMM is a six-dimensional first-order stochastic differential equation (SDE) with four parameters of interest $\theta = [C, \mu, \sigma, g]^T$ and the following mathematical expression:
\begin{equation}
\begin{split}
\mathrm{d}Y_0(t) \ &= \ Y_3(t)\mathrm{d}t \\
\mathrm{d}Y_1(t) \ &= \ Y_4(t)\mathrm{d}t \\
\mathrm{d}Y_2(t) \ &= \ Y_5(t)\mathrm{d}t \\
\mathrm{d}Y_3(t) \ &= \ \bigg[ Aa \Big[  \text{sigm}\Big(Y_1(t) - Y_2(t) \Big)  \Big] - 2aY_3(t) - a^2Y_0(t) \bigg]  \mathrm{d}t +\sigma_3 \mathrm{d}W_3(t)\\
\mathrm{d}Y_4(t) \ &= \ \bigg[ Aa \Big[ \mu + C_2\text{sigm}\Big(C_1 Y_0(t) \Big) \Big] - 2aY_4(t) - a^2Y_1(t) \bigg]  \mathrm{d}t +\sigma_4 \mathrm{d}W_4(t)\\
\mathrm{d}Y_5(t) \ &= \ \bigg[ Bb \Big[ C_4\text{sigm}\Big(C_3 Y_0(t) \Big) \Big] - 2bY_5(t) - b^2Y_2(t)  \bigg]  \mathrm{d}t +\sigma_5 \mathrm{d}W_5(t)\\
\end{split}
\end{equation}
The parameters $C_i$ are related via $C_1 = C, C_2 = 0.8 C, C_3 = C_4 = 0.25 C$, $\sigma_4 = \sigma$, $W_i$ are Wiener processes, and all other (hyper)parameters are fixed and chosen according to reasonable values that have been reported before (see Appendix~\ref{app:real-data} or \citet{ableidinger2017stochastic} for parameter values). The measured signal of an EEG recording is then the scaled difference
\begin{equation}
\begin{split}
y(t) = 10^{g/10}\left( y_1(t) - y_2(t) \right)
\end{split}
\end{equation}
As prior distributions over the parameters $\theta$, we follow previous work \citep{linhart2023lc2st} and define
\begin{equation}
\begin{split}
C &\sim \text{Uniform}(10, 250) \\
\mu &\sim \text{Uniform}(50, 500) \\
\sigma &\sim \text{Uniform}(100, 5000) \\
g &\sim \text{Uniform}(-20, 20) \\
\end{split}
\end{equation}
In the SDE model, parameter $C$ describes the degree of connectivity between populations of excitatory and inhibitory neurons. Parameter $g$ is a gain factor that relates a simulated trajectory from a neural mass model to the readout of an EEG. The other two parameters $\mu$ and $\sigma$ can be treated as nuisance parameters and we consider them of little inferential interest here. They represent statistical properties of incoming oscillations and impact the amplitude of the signal $Y$ (see also \citet{ableidinger2017stochastic,rodrigues2021hnpe}). 

To generate synthetic samples from the simulator function, which is required to train a SBI model, we utilize the {Python} package \texttt{jrnmm} \citep{dirmeier2024jrnmm}.

\begin{figure}[h!t]
    \centering
    \includegraphics[width=0.75\textwidth]{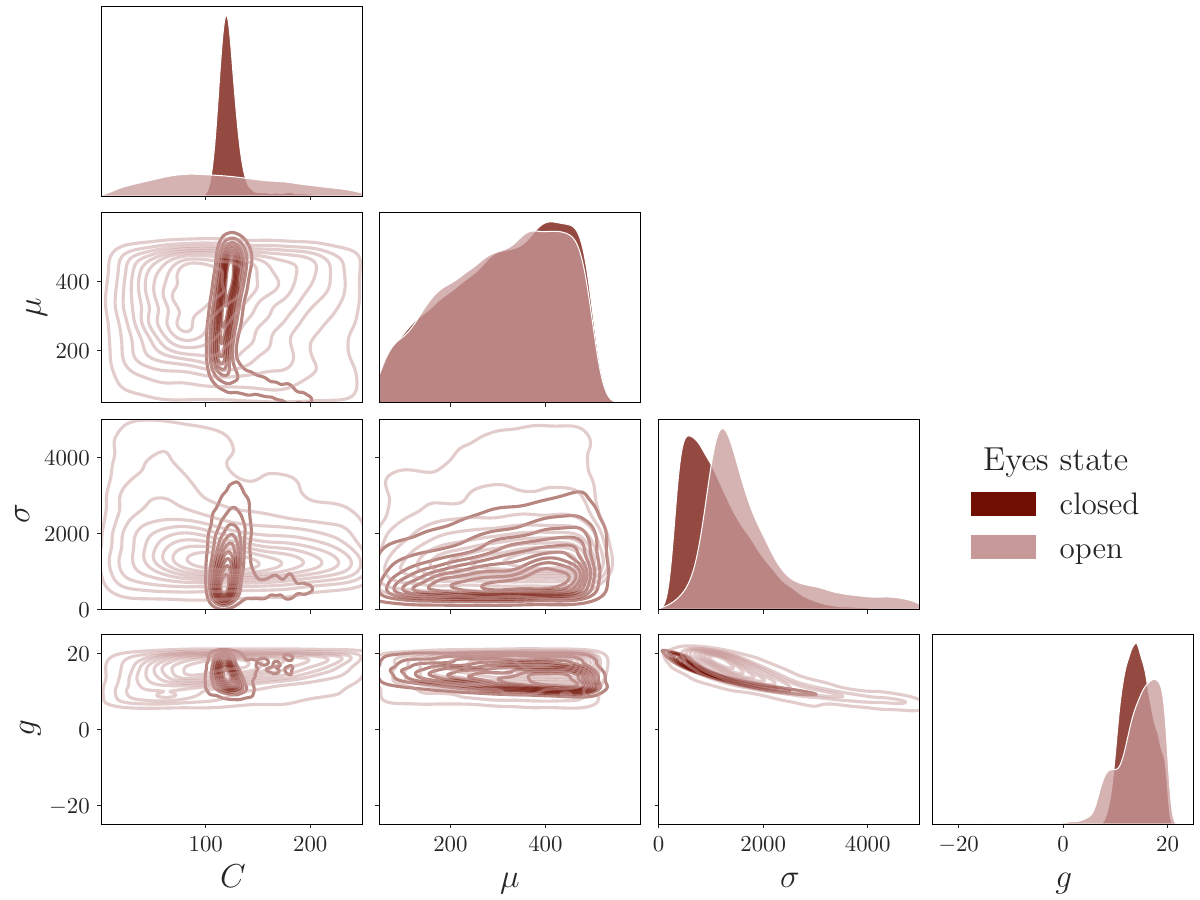}
    \caption{Posterior distributions of the four parameters of the JRNMM model with closed and open eyes. The parameters of primary interest, $C$ and $g$, are well identified when alpha waves are supposedly increased (eyes closed).}
    \label{fig:eeg-posteriors}
\end{figure}

\subsection{Posterior inference}

To learn the conditional density $\pi(\theta | y)$ given EEG measurements, we choose to use the amortized posterior estimator APT. The method constructs a normalizing flow model to directly approximate the intractable posterior distribution (see Table~\ref{tab:all-methods}). The model can then be used to sample from $\pi(\theta | y_\text{closed})$ and $\pi(\theta | y_\text{open})$, using observations $y_\text{closed}$ and $y_\text{open}$ of subjects with closed and open eyes, respectively. We use a masked autoregressive flow with $10$ flow layers each having two hidden layers of $64$ neurons each.

Instead of using the raw EEG data for inference, we compute a set of summary statistics by evaluating the power spectral density (PSD) of the raw EEG signals in frequency bins between $0Hz$ and $64Hz$ (Figure~\ref{fig:eeg-periodogram}; \citet{buckwar2020spectral}, \citet{rodrigues2021hnpe}). Concretely, we first simulate a sample $\{(y_n, \theta_n)\}_{n=1}^{N}$ of size $N=100\ 000$ from the prior and simulator models, then evaluate the PSD of each signal yielding a data set $\{s_n, \theta_n)\}_{n=1}^{N}$ where $s_n = \text{psd}(y_n)$ is the PSD of a signal $y_n$, and then train a {APT} model approximating
\begin{align}
q_\phi(s, \theta) \approx \pi(\theta | y)
\end{align}
Given the trained model, we can simply sample from $q_\phi(s_\text{closed}, \theta)$ and $q_\phi(s_\text{open}, \theta)$ via rejection sampling (see Appendix~\ref{app:real-data} for details such as the transcription into {Python} code).    

The results of the inferences are visualized in Figure~\ref{fig:eeg-posteriors}. Intriguingly, the trained posterior model is able to distinguish the two conditions well. While the nuisance parameters $\mu$ and $\sigma$ are only weakly identified in both eyes states, the neuronal connectivity parameter $C$ can be clearly located in the closed-eyes EEG measurements, i.e., the condition with increased alpha waves. Moreover, in both conditions the gain parameter $g$ expectedly overlaps and can be well identified meaning that the posterior model is able to relate the JRNMM simulation to the amplitude of the raw EEG signals.

\section{Conclusion} 
\label{sec:summary}
We presented \texttt{sbijax}, a {Python} library for simulation-based inference. \texttt{sbijax} implements recent algorithms from the neural SBI and approximate Bayesian computation literature using a light-weight, object-oriented programming interface. It utilizes \texttt{JAX} for automatic differentiation and high-performance computing on either CPU, GPU or even TPU, and integrates several packages from the \texttt{JAX}-verse for neural network training, MCMC sampling of probability distributions. Additionally, by representing posterior distributions as \texttt{InferenceData} objects, \texttt{sbijax} allows seamless integration with the functionality of \texttt{ArviZ}. 

Recently, simulation-based inference has gained significant traction and a multitude of novel methodology has been proposed, yet no high-quality implementations and packages are available which would allow practitioners to use those algorithms. \texttt{sbijax} is the first library available that offers high-quality implementations of state-of-the-art algorithms such as CMPE or NASS. We believe \texttt{sbijax} offers a valuable contribution to the community with a simple learning curve that makes it applicable to users from various fields.

Future work on \texttt{sbijax} will aim to include implementations of the most recent algorithms in SBI, e.g., \citet{h2024quantile,gloeckler2024allinone,sharrock2024sequential}, model diagnostics, such as prior predictive checks \citep{gabry2019visualization} and discriminative calibration \citep{yao2023discriminative}, methods to assess distributional similarity like local classifier two-sample tests \citep{linhart2023lc2st} or H-divergences \citep{zhao2022comparing}, and improved integration with tools such as \texttt{ArviZ}.

\section*{Acknowledgments}
This research was supported by the Swiss National Science Foundation (Grant No. $200021\_208249$).

\newpage
\printbibliography

@inproceedings{
    sharrock2024sequential,
    title={Sequential Neural Score Estimation: Likelihood-Free Inference with Conditional Score Based Diffusion Models},
    author={Louis Sharrock and Jack Simons and Song Liu and Mark Beaumont},
    booktitle={Forty-first International Conference on Machine Learning},
    year={2024}
}

@inproceedings{
    gloeckler2024allinone,
    title={All-in-one simulation-based inference},
    author={Manuel Gloeckler and Michael Deistler and Christian Dietrich Weilbach and Frank Wood and Jakob H. Macke},
    booktitle={Forty-first International Conference on Machine Learning},
    year={2024}
}

@inproceedings{dirmeier2025simulationbased,
    title={Simulation-based Inference for High-dimensional Data using Surjective Sequential Neural Likelihood Estimation},
    author={Simon Dirmeier and Carlo Albert and Fernando Perez-Cruz},
    booktitle={The 41st Conference on Uncertainty in Artificial Intelligence},
    year={2025}
}

@article{dirmeier2024surjectors,
    author = {Simon Dirmeier},
    title = {Surjectors: surjection layers for density estimation with normalizing flows},
    year = {2024},
    journal = {Journal of Open Source Software},
    publisher = {The Open Journal},
    volume = {9},
    number = {94},
    pages = {6188}
}

@InProceedings{papamakarios2019sequential,
  title = 	 {Sequential Neural Likelihood: Fast Likelihood-free Inference with Autoregressive Flows},
  author =       {Papamakarios, George and Sterratt, David and Murray, Iain},
  booktitle =  {Proceedings of the 22nd International Conference on Artificial Intelligence and Statistics},
  year = 	 {2019}
}

@article{papamakarios2021normalizing,
  title={Normalizing flows for probabilistic modeling and inference},
  author={Papamakarios, George and Nalisnick, Eric and Rezende, Danilo Jimenez and Mohamed, Shakir and Lakshminarayanan, Balaji},
  journal={The Journal of Machine Learning Research},
  volume={22},
  number={1},
  pages={2617--2680},
  year={2021},
  publisher={JMLRORG}
}

@inproceedings{
    wildberger2023flow,
    title={Flow Matching for Scalable Simulation-Based Inference},
    author={Jonas Bernhard Wildberger and Maximilian Dax and Simon Buchholz and Stephen R Green and Jakob H. Macke and Bernhard Sch{\"o}lkopf},
    booktitle={Advances in Neural Information Processing Systems},
    year={2023},
}

@inproceedings{
    chen2021neural,
    title={Neural Approximate Sufficient Statistics for Implicit Models},
    author={Yanzhi Chen and Dinghuai Zhang and Michael U. Gutmann and Aaron Courville and Zhanxing Zhu},
    booktitle={International Conference on Learning Representations},
    year={2021}
}

@article{schmitt2023consistency,
  title={Consistency Models for Scalable and Fast Simulation-Based Inference},
  author={Schmitt, Marvin and Pratz, Valentin and K{\"o}the, Ullrich and B{\"u}rkner, Paul-Christian and Radev, Stefan T},
  journal={arXiv preprint arXiv:2312.05440},
  year={2023}
}

@inproceedings{chen2023learning,
  title={Is learning summary statistics necessary for likelihood-free inference?},
  author={Chen, Yanzhi and Gutmann, Michael U and Weller, Adrian},
  booktitle={Proceedings of the 40th International Conference on Machine
Learning},  
  year={2023}
}

@inproceedings{greenberg2019automatic,
  title = 	 {Automatic Posterior Transformation for Likelihood-Free Inference},
  author =       {Greenberg, David and Nonnenmacher, Marcel and Macke, Jakob},
  booktitle = 	 {Proceedings of the 36th International Conference on Machine Learning},
  year = 	 {2019},
}

@inproceedings{miller2022contrastive,
    title={Contrastive Neural Ratio Estimation},
    author={Benjamin Kurt Miller and Christoph Weniger and Patrick Forr{\'e}},
    booktitle={Advances in Neural Information Processing Systems},
    year={2022},
}

@software{deepmind2020jax,
  title = {The {D}eep{M}ind {JAX} {E}cosystem},
  author = {DeepMind and Babuschkin, Igor and Baumli, Kate and Bell, Alison and Bhupatiraju, Surya and Bruce, Jake and Buchlovsky, Peter and Budden, David and Cai, Trevor and Clark, Aidan and Danihelka, Ivo and Dedieu, Antoine and Fantacci, Claudio and Godwin, Jonathan and Jones, Chris and Hemsley, Ross and Hennigan, Tom and Hessel, Matteo and Hou, Shaobo and Kapturowski, Steven and Keck, Thomas and Kemaev, Iurii and King, Michael and Kunesch, Markus and Martens, Lena and Merzic, Hamza and Mikulik, Vladimir and Norman, Tamara and Papamakarios, George and Quan, John and Ring, Roman and Ruiz, Francisco and Sanchez, Alvaro and Sartran, Laurent and Schneider, Rosalia and Sezener, Eren and Spencer, Stephen and Srinivasan, Srivatsan and Stanojevi\'{c}, Milo\v{s} and Stokowiec, Wojciech and Wang, Luyu and Zhou, Guangyao and Viola, Fabio},
  url = {http://github.com/deepmind},
  year = {2020},
}

@software{jax2018github,
  author = {James Bradbury and Roy Frostig and Peter Hawkins and Matthew James Johnson and Chris Leary and Dougal Maclaurin and George Necula and Adam Paszke and Jake Vander{P}las and Skye Wanderman-{M}ilne and Qiao Zhang},
  title = {{JAX}: composable transformations of {P}ython+{N}um{P}y programs},
  url = {http://github.com/google/jax},
  version = {0.3.13},
  year = {2018},
}

@article{cabezas2024blackjax,
  title={BlackJAX: Composable {B}ayesian inference in {JAX}},
  author={Alberto Cabezas and Adrien Corenflos and Junpeng Lao and Rémi Louf},
  year={2024},
  journal={arXiv preprint arXiv:2402.10797},
}

@article{dillon2017tensorflow,
  title={Tensorflow distributions},
  author={Dillon, Joshua V and Langmore, Ian and Tran, Dustin and Brevdo, Eugene and Vasudevan, Srinivas and Moore, Dave and Patton, Brian and Alemi, Alex and Hoffman, Matt and Saurous, Rif A},
  journal={arXiv preprint arXiv:1711.10604},
  year={2017}
}

@article{kumar2019arviz, 
  year = {2019},  
  volume = {4},
  number = {33},
  pages = {1143},
  author = {Ravin Kumar and Colin Carroll and Ari Hartikainen and Osvaldo Martin},
  title = {ArviZ a unified library for exploratory analysis of Bayesian models in Python},
  journal = {Journal of Open Source Software}
}

@software{haiku2020github,
  author = {Tom Hennigan and Trevor Cai and Tamara Norman and Lena Martens and Igor Babuschkin},
  title = {{H}aiku: {S}onnet for {JAX}},
  url = {http://github.com/deepmind/dm-haiku},
  version = {0.0.10},
  year = {2020},
}

@software{dirmeier2024jrnmm,
  author  = {Simon Dirmeier},
  title   = {jrnmm: The Jansen-Rit neural mass model SDE in JAX},
  year    = {2024},
  version = {0.1.0.post1},
  url = {http://github.com/dirmeier/jrnmm}
}

@article{bishop1994mixture,
  title={Mixture density networks},
  author={Bishop, Christopher M},
  year={1994},
  publisher={Aston University}
}

@article{hoffman2014no,
  title={The No-U-Turn sampler: adaptively setting path lengths in Hamiltonian Monte Carlo.},
  author={Hoffman, Matthew D and Gelman, Andrew and others},
  journal={Journal of Machine Learning Research},
  volume={15},
  number={1},
  pages={1593--1623},
  year={2014}
}

@article{neal2003slice,
  title={Slice sampling},
  author={Neal, Radford M},
  journal={The Annals of Statistics},
  volume={31},
  number={3},
  pages={705--767},
  year={2003}
}

@article{blei2017variational,
  title={Variational inference: A review for statisticians},
  author={Blei, David M and Kucukelbir, Alp and McAuliffe, Jon D},
  journal={Journal of the American statistical Association},
  volume={112},
  number={518},
  pages={859--877},
  year={2017},
  publisher={Taylor \& Francis}
}

@inproceedings{lipman2023flow,
    title={Flow Matching for Generative Modeling},
    author={Yaron Lipman and Ricky T Q Chen and Heli Ben-Hamu and Maximilian Nickel and Matthew Le},
    booktitle={The Eleventh International Conference on Learning Representations},
    year={2023}
}

@inproceedings{chen2018neurao,
    author = {Chen, Ricky T Q and Rubanova, Yulia and Bettencourt, Jesse and Duvenaud, David K},
    booktitle = {Advances in Neural Information Processing Systems},
    editor = {S. Bengio and H. Wallach and H. Larochelle and K. Grauman and N. Cesa-Bianchi and R. Garnett},     
    title = {Neural Ordinary Differential Equations},     
    year = {2018}
}

@book{sugiyama2012density,
  title={Density ratio estimation in machine learning},
  author={Sugiyama, Masashi and Suzuki, Taiji and Kanamori, Takafumi},
  year={2012},
  publisher={Cambridge University Press}
}

@inproceedings{paszke2019pytorch,
 author = {Paszke, Adam and Gross, Sam and Massa, Francisco and Lerer, Adam and Bradbury, James and Chanan, Gregory and Killeen, Trevor and Lin, Zeming and Gimelshein, Natalia and Antiga, Luca and Desmaison, Alban and Kopf, Andreas and Yang, Edward and DeVito, Zachary and Raison, Martin and Tejani, Alykhan and Chilamkurthy, Sasank and Steiner, Benoit and Fang, Lu and Bai, Junjie and Chintala, Soumith},
 booktitle = {Advances in Neural Information Processing Systems},
 title = {PyTorch: An Imperative Style, High-Performance Deep Learning Library}, 
 year = {2019},
}

@article{tejero-cantero2020sbi,  
    year = {2020}, 
    publisher = {The Open Journal}, 
    volume = {5}, number = {52}, 
    pages = {2505}, 
    author = {Alvaro Tejero-Cantero and Jan Boelts and Michael Deistler and Jan-Matthis Lueckmann and Conor Durkan and Pedro J. Gonçalves and David S. Greenberg and Jakob H. Macke}, 
    title = {sbi: A toolkit for simulation-based inference},
    journal = {Journal of Open Source Software} 
}

@book{sisson2018handbook,
  title={Handbook of approximate Bayesian computation},
  author={Sisson, Scott A and Fan, Yanan and Beaumont, Mark},
  year={2018},
  publisher={CRC Press}
}

@book{brooks2011mcmc,
  author = {Brooks, Steve and Gelman, Andrew and Jones, Galin and Meng, Xiao-Li},
  publisher = {CRC press},  
  title = {Handbook of Markov Chain Monte Carlo},
  year = {2011},
}

@article{albert2015simulated,
  title={A simulated annealing approach to approximate {B}ayes computations},
  author={Albert, Carlo and K{\"u}nsch, Hans R and Scheidegger, Andreas},
  journal={Statistics and Computing},
  volume={25},
  pages={1217--1232},
  year={2015},
  publisher={Springer}
}

@article{beaumont2009adaptive,
  title={Adaptive approximate Bayesian computation},
  author={Beaumont, Mark A and Cornuet, Jean-Marie and Marin, Jean-Michel and Robert, Christian P},
  journal={Biometrika},
  volume={96},
  number={4},
  pages={983--990},
  year={2009},
  publisher={Oxford University Press}
}

@article{del2012adaptive,
  title={An adaptive sequential Monte Carlo method for approximate Bayesian computation},
  author={Del Moral, Pierre and Doucet, Arnaud and Jasra, Ajay},
  journal={Statistics and Computing},
  volume={22},
  pages={1009--1020},
  year={2012},
  publisher={Springer}
}

@article{vehtari2021split,
    author = {Aki Vehtari and Andrew Gelman and Daniel Simpson and Bob Carpenter and Paul-Christian B{\"u}rkner},
    title = {{Rank-Normalization, Folding, and Localization: An Improved $\widehat{R}$ for Assessing Convergence of MCMC (with Discussion)}},
    volume = {16},
    journal = {Bayesian Analysis},
    number = {2},
    publisher = {International Society for Bayesian Analysis},
    pages = {667 -- 718},
    year = {2021},
}

@article{hunter2007mat,
  Author    = {Hunter, J. D.},
  Title     = {Matplotlib: A 2D graphics environment},
  Journal   = {Computing in Science \& Engineering},
  Volume    = {9},
  Number    = {3},
  Pages     = {90--95},
  publisher = {IEEE COMPUTER SOC},
  year = {2006}
}

@inproceedings{he2016deep,
    author = {He, Kaiming and Zhang, Xiangyu and Ren, Shaoqing and Sun, Jian},
    title = {Deep Residual Learning for Image Recognition},
    booktitle = {Proceedings of the IEEE Conference on Computer Vision and Pattern Recognition (CVPR)},    
    year = {2016}
}

@inproceedings{durkan2019neural,
  title={Neural spline flows},
  author={Durkan, Conor and Bekasov, Artur and Murray, Iain and Papamakarios, George},
  booktitle={Advances in Neural Information Processing Systems},  
  year={2019}
}

@inproceedings{papamakarios2017masked,
  title={Masked autoregressive flow for density estimation},
  author={Papamakarios, George and Pavlakou, Theo and Murray, Iain},
  booktitle={Advances in Neural Information Processing Systems},  
  year={2017}
}

@inproceedings{dinh2017density,
    title={Density estimation using Real {NVP}},
    author={Laurent Dinh and Jascha Sohl-Dickstein and Samy Bengio},
    booktitle={International Conference on Learning Representations},
    year={2017}
}

@inproceedings{song2023consistency,
    title={Consistency Models}, 
    author={Yang Song and Prafulla Dhariwal and Mark Chen and Ilya Sutskever},
    year={2023},
    booktitle={Proceedings of the 40th International Conference on Machine Learning},      
}

@article{cranmer2020frontier,
  title={The frontier of simulation-based inference},
  author={Cranmer, Kyle and Brehmer, Johann and Louppe, Gilles},
  journal={Proceedings of the National Academy of Sciences},
  volume={117},
  number={48},
  pages={30055--30062},
  year={2020},
  publisher={National Acadademy of Sciences}
}

@article{harris2020array,
  title={Array programming with NumPy},
  author={Harris, Charles R and Millman, K Jarrod and Van Der Walt, St{\'e}fan J and Gommers, Ralf and Virtanen, Pauli and Cournapeau, David and Wieser, Eric and Taylor, Julian and Berg, Sebastian and Smith, Nathaniel J and others},
  journal={Nature},
  volume={585},
  number={7825},
  pages={357--362},
  year={2020},
  publisher={Nature Publishing Group UK London}
}

@misc{gabry2024bayesplot,
  title = {bayesplot: Plotting for Bayesian Models},
  author = {Jonah Gabry and Tristan Mahr},
  year = {2024},
  note = {R package version 1.11.0},
  url = {https://mc-stan.org/bayesplot/},
}

@article{gabry2019visualization,
  title = {Visualization in Bayesian workflow},
  author = {Jonah Gabry and Daniel Simpson and Aki Vehtari and Michael Betancourt and Andrew Gelman},
  year = {2019},
  journal = {J. R. Stat. Soc. A},
  volume = {182},
  issue = {2},
  pages = {389-402},
}

@inproceedings{hermans2020likelihood,
  title = 	 {Likelihood-free {MCMC} with Amortized Approximate Ratio Estimators},
  author =       {Hermans, Joeri and Begy, Volodimir and Louppe, Gilles},
  booktitle = 	 {Proceedings of the 37th International Conference on Machine Learning},  
  year = 	 {2020},
}

@article{hoyer2017xarray,
  title     = {xarray: {N-D} labeled arrays and datasets in {Python}},
  author    = {Hoyer, S. and J. Hamman},
  journal   = {Journal of Open Research Software},
  volume    = {5},
  number    = {1},
  year      = {2017}
}

@inproceedings{yao2023discriminative,
 author = {Yao, Yuling and Domke, Justin},
 booktitle = {Advances in Neural Information Processing Systems}, 
 title = {Discriminative Calibration: Check Bayesian Computation from Simulations and Flexible Classifier},
 year = {2023}
}

@inproceedings{linhart2023lc2st,
 author = {Linhart, Julia and Gramfort, Alexandre and Rodrigues, Pedro},
 booktitle = {Advances in Neural Information Processing Systems},
 title = {L-C2ST: Local Diagnostics for Posterior Approximations in Simulation-Based Inference}, 
 year = {2023}
}

@inproceedings{zhao2022comparing,
  title={Comparing Distributions by Measuring Differences that Affect Decision Making},
  author={Shengjia Zhao and Abhishek Sinha and Yutong He and Aidan Perreault and Jiaming Song and Stefano Ermon},
  booktitle={International Conference on Learning Representations},
  year={2022}
}

@article{sisson2007sequential,
  title={Sequential Monte Carlo without likelihoods},
  author={Sisson, Scott A and Fan, Yanan and Tanaka, Mark M},
  journal={Proceedings of the National Academy of Sciences},
  volume={104},
  number={6},
  pages={1760--1765},
  year={2007},
}

@article{charbonneau2005fluctuations,
  title={Fluctuations in {B}abcock-{L}eighton dynamos. I. period doubling and transition to chaos},
  author={Charbonneau, Paul and St-Jean, C{\'e}dric and Zacharias, Pia},
  journal={The Astrophysical Journal},
  volume={619},
  number={1},
  pages={613},
  year={2005},
  publisher={IOP Publishing}
}

@inproceedings{
    liu2023flow,
    title={Flow Straight and Fast: Learning to Generate and Transfer Data with Rectified Flow},
    author={Xingchao Liu and Chengyue Gong and qiang liu},
    booktitle={The Eleventh International Conference on Learning Representations },
    year={2023}
}

@article{phan2019composable,
  title={Composable effects for flexible and accelerated probabilistic programming in NumPyro},
  author={Phan, Du and Pradhan, Neeraj and Jankowiak, Martin},
  journal={arXiv preprint arXiv:1912.11554},
  year={2019}
}

@InProceedings{h2024quantile,
  title = 	 {Simulation-Based Inference with Quantile Regression},
  author =       {Jia, He},
  booktitle = 	 {Proceedings of the 41st International Conference on Machine Learning},  
  year = 	 {2024},
}

@article{tierney1994markov,
  title={Markov chains for exploring posterior distributions},
  author={Tierney, Luke},
  journal={The Annals of Statistics},
  pages={1701--1728},
  year={1994}
}

@book{compbayes2019,     
    title={Computational Bayesian Statistics: An Introduction}, 
    publisher={Cambridge University Press},
    author={Amaral Turkman, M. Antónia and Paulino, Carlos Daniel and Müller, Peter}, 
    year={2019}
}

@article{schaelte2022,
	title        = {pyABC: Efficient and robust easy-to-use approximate Bayesian computation},
	author       = {Yannik Schälte and Emmanuel Klinger and Emad Alamoudi and Jan Hasenauer},
	year         = 2022,
	journal      = {Journal of Open Source Software},
	publisher    = {The Open Journal},
	volume       = 7,
	number       = 74,
	pages        = 4304,
}

@article{dutta2021abcpy,
    title={ABCpy: A High-Performance Computing Perspective to Approximate Bayesian Computation},
    volume={100},
    number={7},
    journal={Journal of Statistical Software},
    author={Dutta, Ritabrata and Schoengens, Marcel and Pacchiardi, Lorenzo and Ummadisingu, Avinash and Widmer, Nicole and Künzli, Pierre and Onnela, Jukka-Pekka and Mira, Antonietta},
    year={2021},
    pages={1–38}
}

@article{approxbayescomp,
  title={Approximate Bayesian Computations to fit and compare insurance loss models},
  author={Goffard, Pierre-Olivier and Laub, Patrick J},
  journal={Insurance: Mathematics and Economics},
  volume={100},
  pages={350--371},
  year={2021}
}

@article{rousset2017summary,
  title={The summary-likelihood method and its implementation in the Infusion package},
  author={Rousset, Francois and Gouy, Alexandre and Martinez-Almoyna, Camille and Courtiol, Alexandre},
  journal={Molecular ecology resources},
  volume={17},
  number={1},
  pages={110--119},
  year={2017},
  publisher={Wiley Online Library}
}

@article{rodrigues2021hnpe,
  title={HNPE: Leveraging global parameters for neural posterior estimation},
  author={Rodrigues, Pedro and Moreau, Thomas and Louppe, Gilles and Gramfort, Alexandre},
  journal={Advances in Neural Information Processing Systems},
  volume={34},
  pages={13432--13443},
  year={2021}
}

@article{ableidinger2017stochastic,
	author = {Ableidinger, Markus and Buckwar, Evelyn and Hinterleitner, Harald},	
	journal = {The Journal of Mathematical Neuroscience},
	number = {1},
	pages = {8},
	title = {A Stochastic Version of the {J}ansen and {R}it Neural Mass Model: Analysis and Numerics},	
	volume = {7},
	year = {2017}
}

@article{buckwar2020spectral,
  title={Spectral density-based and measure-preserving {ABC} for partially observed diffusion processes. An illustration on {H}amiltonian {SDE}s},
  author={Buckwar, Evelyn and Tamborrino, Massimiliano and Tubikanec, Irene},
  journal={Statistics and Computing},
  volume={30},
  number={3},
  pages={627--648},
  year={2020}
}

@dataset{cattan2018,
  author       = {Grégoire Cattan and
                  Pedro L. C. Rodrigues and
                  Marco Congedo},
  title        = {EEG Alpha Waves dataset},
  year         = 2018,
  publisher    = {Zenodo},
  doi          = {10.5281/zenodo.2348892}
}

@article{virtanen2020scipy,
  title={SciPy 1.0: fundamental algorithms for scientific computing in Python},
  author={Virtanen, Pauli and Gommers, Ralf and Oliphant, Travis E and Haberland, Matt and Reddy, Tyler and Cournapeau, David and Burovski, Evgeni and Peterson, Pearu and Weckesser, Warren and Bright, Jonathan and others},
  journal={Nature methods},
  volume={17},
  number={3},
  pages={261--272},
  year={2020},
  publisher={Nature Publishing Group}
}

\newpage
\appendix
\section{Experimental details}
\label{app:experimental-details}
All experiments have been conducted on a Macbook Pro M1 (Ventura 13.0.1, 8 Core CPU, 16 GB RAM) using Python 3.11.7. The exact versions of each library can be found in the supplementary material. Since \texttt{sbijax} and its dependencies use \texttt{JAX} extensively (which uses 32bit floating point arithmetic and SIMD procedures for computational speed), we cannot guarantee exact replicability. 
Generally, using the same OS, CPU and the same library dependencies, only very minor differences should be expected, for instance, when running MCMC chains or conducting neural network training. On different systems, the differences might be slightly bigger.

\section{Additional material}
\begin{figure}[h]
\centering
\includegraphics[width=.8\textwidth]{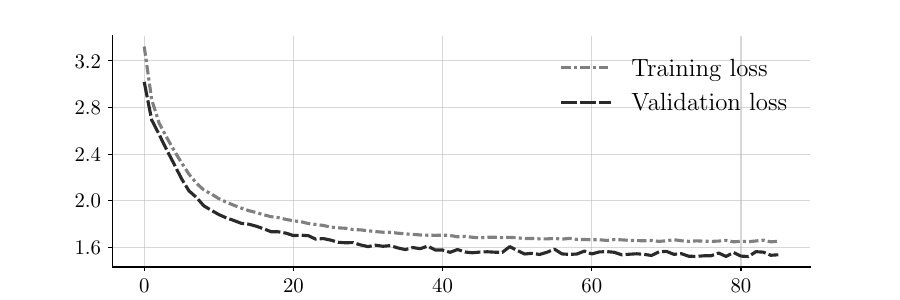}
\caption{Training and validation loss for bivariate Gaussian example. The training of the neural network converged in this example and was stopped early after roughly $130$ episodes, because there were only insignificant improvements on the validation set.}
\label{fig:bivariate_gaussian-losses}
\end{figure}

\section{Additional material for the SLCP example}
\label{app:additional-material-for-slcp}

Below, we provide additional material for Section~\ref{sec:examples}. The generative model of the simple-likelihood-complex-posterior (SLCP) example can be expressed in following {Python} code.

\begin{small}
\begin{verbatim}
def prior_fn():
    prior = tfd.JointDistributionNamed(dict(
        theta=tfd.Uniform(jnp.full(5, -3.0), jnp.full(5, 3.0))
    ), batch_ndims=0)
    return prior
    
def simulator_fn(seed, theta):
    theta = theta["theta"]    
    theta = theta[:, None, :]
    us_key, noise_key = jr.split(seed)

    def _unpack_params(ps):
        m0 = ps[..., [0]]
        m1 = ps[..., [1]]
        s0 = ps[..., [2]] ** 2
        s1 = ps[..., [3]] ** 2
        r = jnp.tanh(ps[..., [4]])
        return m0, m1, s0, s1, r

    m0, m1, s0, s1, r = _unpack_params(theta)
    us = tfd.Normal(0.0, 1.0).sample(
        seed=us_key, 
        sample_shape=(theta.shape[0], theta.shape[1], 4, 2)
    )
    xs = jnp.empty_like(us)
    xs = xs.at[:, :, :, 0].set(s0 * us[:, :, :, 0] + m0)
    y = xs.at[:, :, :, 1].set(
        s1 * (r * us[:, :, :, 0] + jnp.sqrt(1.0 - r**2) * us[:, :, :, 1]) + m1
    )    
    y = y.reshape((*theta.shape[:1], 8))    
    return y
\end{verbatim}
\end{small}

We use the same observation as in \citet{papamakarios2019sequential}, namely:
\begin{equation*}
    y_\text{obs} = \begin{bmatrix}
     -0.9707123 \\
    -2.9461224\\
    -0.4494722\\
    -3.4231849\\
    -0.1328563\\
    -3.3640170\\
    -0.8536759\\
    -2.4271638\\
    \end{bmatrix}
\end{equation*}

\begin{figure}[h]
\centering
\includegraphics[width=1\textwidth]{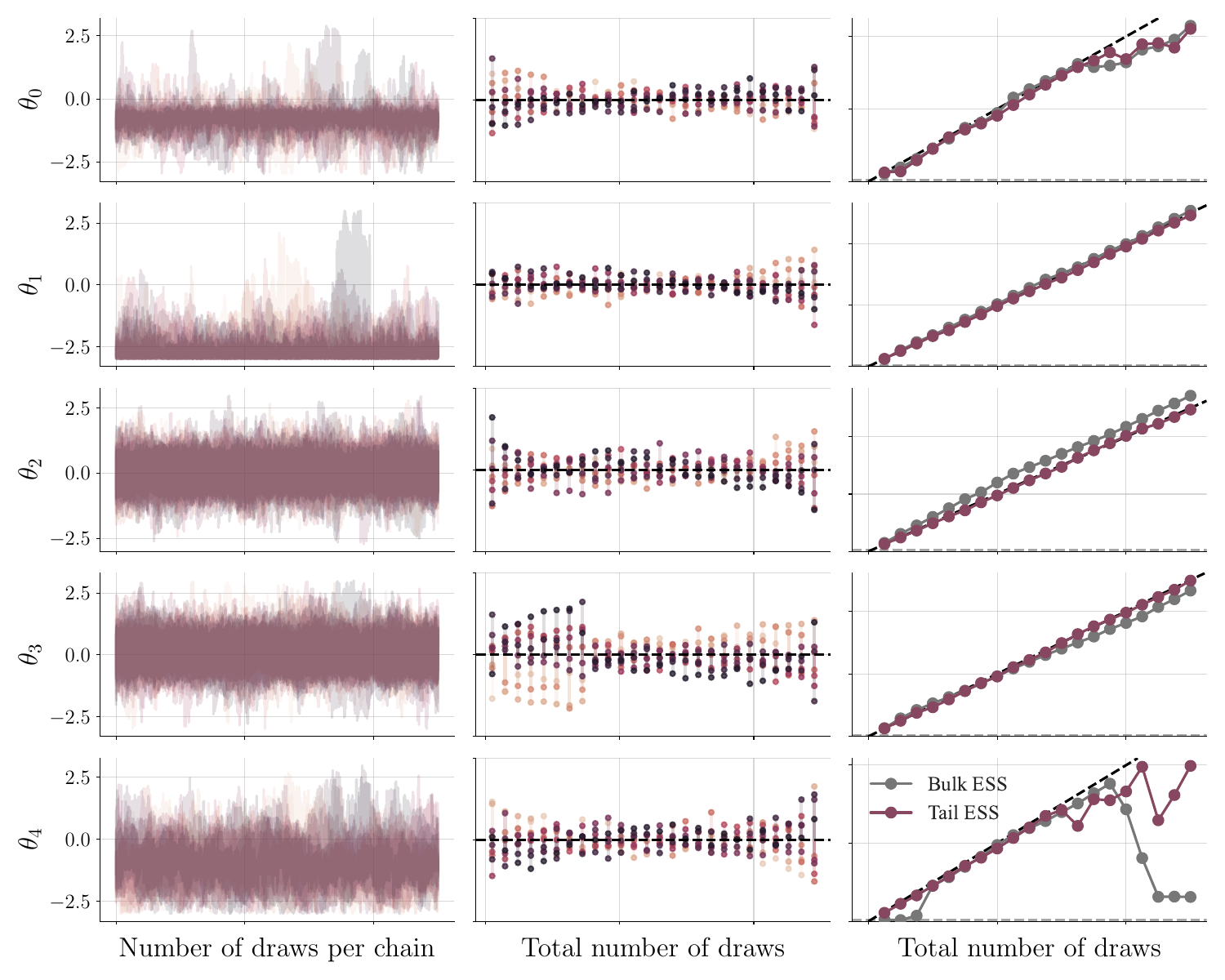}
\caption{MCMC model diagnostics for SLCP model. We show three common MCMC model diagnostics for which \texttt{sbijax} offers functionality for visualization. The left column shows posterior traces, i.e., the values of $\theta$ for each iteration and for each chain (different colors). The column in the middle shows rank statistics for each parameter and chain  (different colors). The right column shows the bulk and tail effective sample sizes
}
\label{fig:slcp-slice-tre-plot}
\end{figure}

\begin{figure}[h]
\centering
\includegraphics[width=0.8\textwidth]{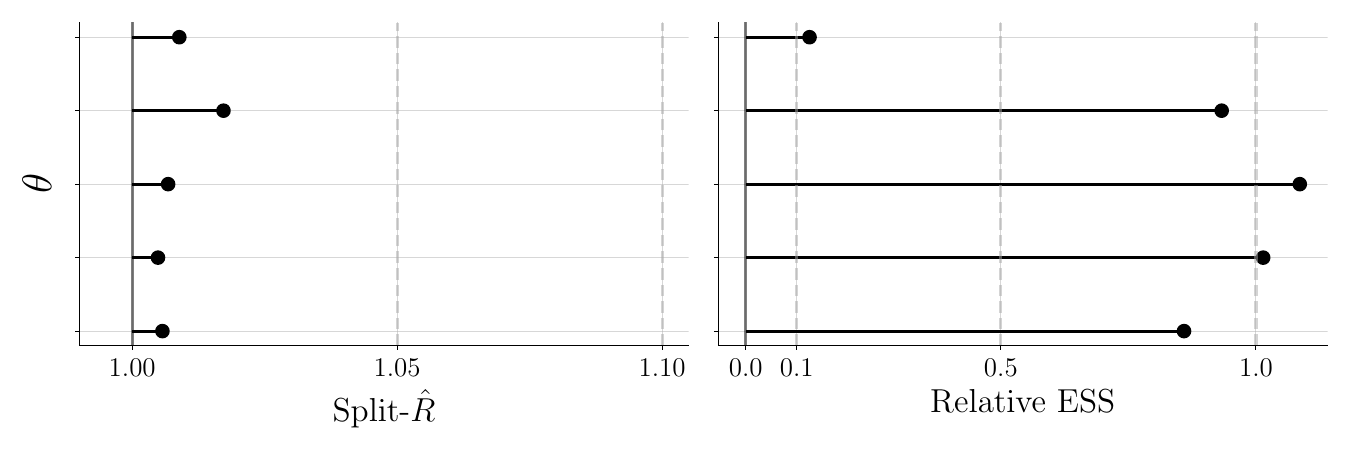}
\caption{MCMC model diagnostics for SLCP model. Both split-$\hat{R}$ and relative ESS look reasonable for all parameters (i.e., they are roughly $1$ for the split-$\hat{R}$ and larger than $0.1$ for the relative ESS).}
\label{fig:slcp-slice-rhat_ress}
\end{figure}

\newpage
\section{Additional algorithm examples}
\label{app:additional-algorithm-examples}
The following sections demonstrate more code examples using different inferential algorithms implemented in \texttt{sbijax}. We demonstrate the algorithms on a simple mixture model which in a similar form has been used as a benchmark example in the SBI literature before \citep{sisson2007sequential,beaumont2009adaptive}.

The statistical model has the form:
\begin{equation}
\begin{split}
\theta &\sim \mathcal{N}_2(0, I)\\
y \mid \theta &\sim 0.5 \ \mathcal{N}_2(\theta, I) + 0.5 \ \mathcal{N}_2(\theta, 0.01 I)
\end{split}
\label{app:eqn-mixture}
\end{equation}
It is a simple mixture where the mixing weights and covariance parameters are fixed and only the mean of the two Gaussians is random. The inferential task is to infer the posterior $\pi(\theta | y_{\text{obs}})$. The transcription of the prior model and the simulator into {Python} code is shown below.

\begin{small}
\begin{verbatim}
def prior_fn():
    prior = tfd.JointDistributionNamed(dict(
        theta=tfd.Normal(jnp.zeros(2), 1)
    ), batch_ndims=0)
    return prior

def simulator_fn(seed, theta):
    mean = theta["theta"].reshape(-1, 2)
    n = mean.shape[0]
    data_key, cat_key = jr.split(seed)
    pi_categories = tfd.Categorical(logits=jnp.zeros(2))
    categories = pi_categories.sample(seed=cat_key, sample_shape=(n,))
    scales = jnp.array([1.0, 0.1])[categories].reshape(-1, 1)
    y = tfd.Normal(mean, scales).sample(seed=data_key)
    return y
\end{verbatim}
\end{small}
As an observation, we arbitrarily choose $y_{\text{obs}} = [-1.0, 1.0]^T$.
\begin{small}
\begin{verbatim}
y_observed = jnp.array([-1.0, 1.0])
\end{verbatim}
\end{small}

\begin{figure}
  \centering
        \begin{subfigure}[b]{0.49\textwidth}
            \centering
            \includegraphics[width=\textwidth]{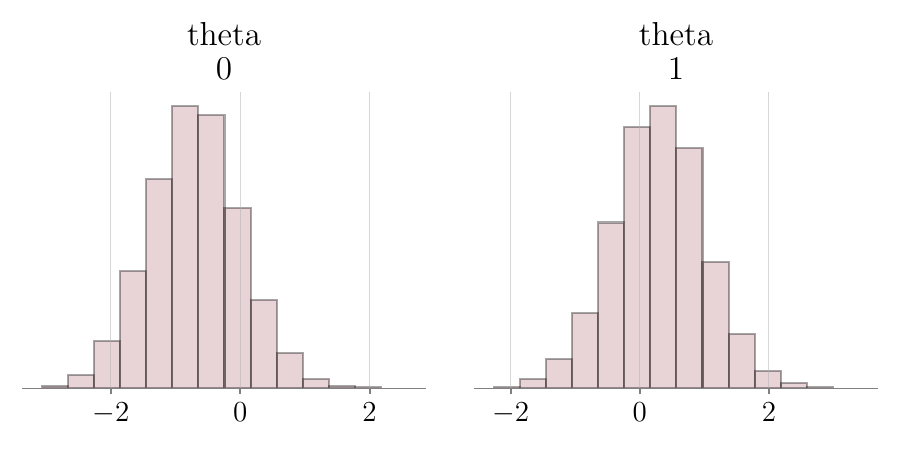}
            \caption{{No-U-Turn sampling posterior.}}      
            \label{app:mixture_model-plots-nuts}
        \end{subfigure}        
        \vskip\baselineskip  
        \begin{subfigure}[b]{0.49\textwidth}   
            \centering 
            \includegraphics[width=\textwidth]{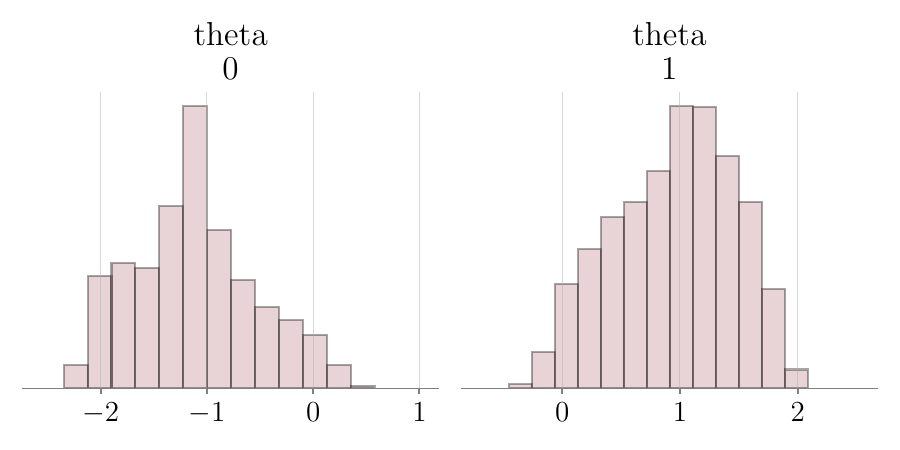}
            \caption{{CMPE posterior.}}         
            \label{app:mixture_model-plots-cmpe}
        \end{subfigure}                       
        \begin{subfigure}[b]{0.49\textwidth}  
            \centering 
            \includegraphics[width=\textwidth]{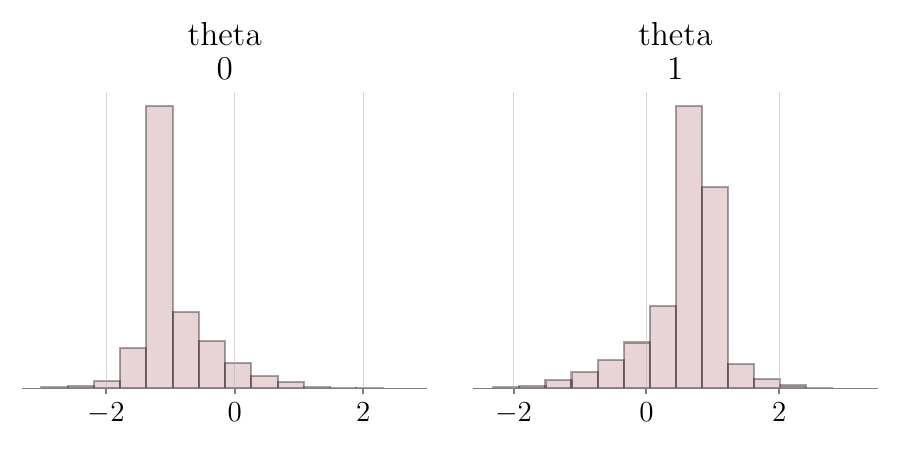}
            \caption{{NRE posterior.}}         
            \label{app:mixture_model-plots-nre}
        \end{subfigure}
        
        \vskip\baselineskip    
        \begin{subfigure}[b]{0.49\textwidth}   
            \centering 
            \includegraphics[width=\textwidth]{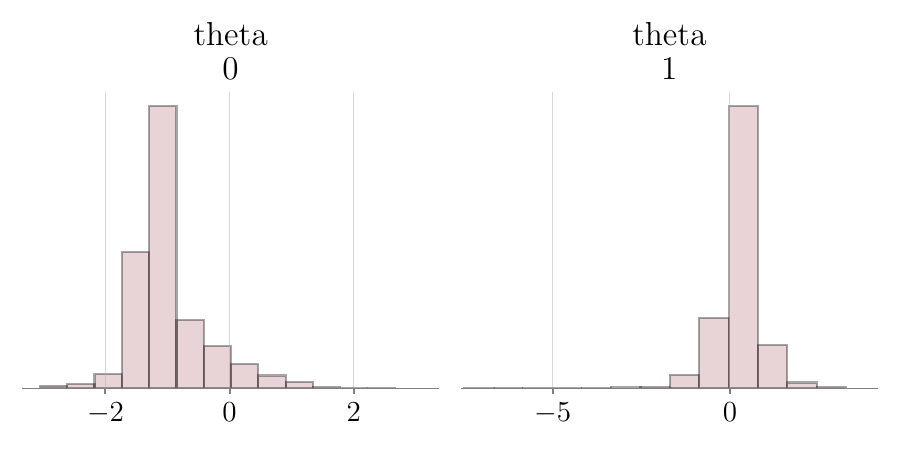}
            \caption{{NPE posterior.}}            
            \label{app:mixture_model-plots-npe}
        \end{subfigure}                    
        \hfill
        \begin{subfigure}[b]{0.49\textwidth}   
            \centering 
            \includegraphics[width=\textwidth]{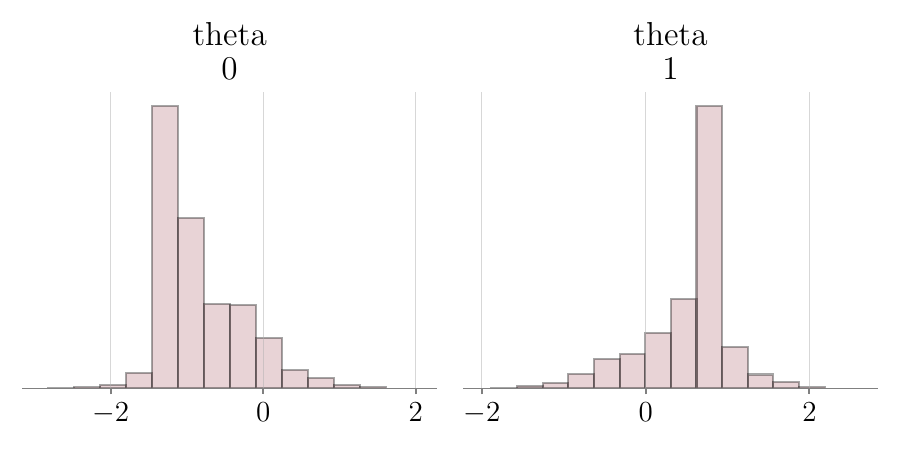}
            \caption{{NLE posterior.}}        
            \label{app:mixture_model-plots-nle}
        \end{subfigure}
        \caption{Posterior distribution of different algorithms on a simple mixture model. The posterior inferred using a No-U-Turn sampler represents the true posterior distribution (up to some MC-error).}
\label{app:mixture_model-plots}
\end{figure}

In the following, we demonstrate \textit{consistency model posterior estimation} (CMPE; \citet{schmitt2023consistency}),
\textit{automatic posterior transformation} (which we here denote as NPE; \citep{greenberg2019automatic}), contrastive neural ratio estimation (NRE; \citet{miller2022contrastive}) and \textit{neural likelihood estimation} (NLE; \citep{papamakarios2019sequential}). We compare the inferred posteriors to a posterior that has been inferred using a No-U-Turn sampler in order to be able to compare the accuracy of the methods. The convergence of the sampler has been diagnosed using the split-$\hat{R}$ statistic, the effective sample size and graphical trace plots. The posterior distributions are visualized in figure~\ref{app:mixture_model-plots}.

\subsection{Consistency model posterior estimation}
CMPE requires a consistency model as neural network architecture for inference. \texttt{sbijax} provides functionality to construct a consistency model using \texttt{make\_cm}, but here we will use the low-level API. We define a simple multi-layer perceptron using the package \texttt{Haiku} and design an neural network architecture as described in the original publication by \citet{schmitt2023consistency}:

\begin{small}
\begin{verbatim}
import haiku as hk

def make_model(dim):
    @hk.transform
    def mlp(method, **kwargs):
        def _c_skip(time):
            return 1 / ((time - 0.001) ** 2 + 1)

        def _c_out(time):
            return 1.0 * (time - 0.001) / jnp.sqrt(1 + time**2)

        def _nn(theta, time, context, **kwargs):
            ins = jnp.concatenate([theta, time, context], axis=-1)
            outs = hk.nets.MLP([64, 64, dim])(ins)
            out_skip = _c_skip(time) * theta + _c_out(time) * outs
            return out_skip

        cm = ConsistencyModel(dim, _nn)
        return cm(method, **kwargs)
    
    return mlp
\end{verbatim}
\end{small}

We can then define the CMPE algorithm, simulate artificial data, fit the data to the model, and then draw samples from the approximate posterior distribution (see figure~\ref{app:mixture_model-plots-cmpe} for a visualization).
\begin{small}
\begin{verbatim}
fns = prior_fn, simulator_fn
model = CMPE(fns, make_model(2))

data, _ = model.simulate_data(jr.PRNGKey(1), n_simulations=10_000)
params, _ = model.fit(jr.PRNGKey(2), data=data)

inference_results, _ = model.sample_posterior(jr.PRNGKey(3), params, y_observed)
\end{verbatim}
\end{small}

\subsection{Neural ratio estimation}
Neural ratio-estimation requires a classifier network to infer the posterior distribution. This makes NRE methods extremely attractive, since, in comparison to, e.g., normalizing flows classifiers are trivial to train. We can use the neural network library \texttt{Haiku} to construct a simple MLP with a single node in the last layer. The code is shown below. 
\begin{small}
\begin{verbatim}
def make_model():
    @hk.without_apply_rng
    @hk.transform
    def mlp(inputs, **kwargs):
        return hk.nets.MLP([64, 64, 1])(inputs)

    return mlp
\end{verbatim}
\end{small}

As all other neural SBI methods, NRE requires artificial data to fit the network. The trained network can then be used to compute likelihood-ratios and use them to sample from the approximate posterior. Per default, NRE uses a No-U-Turn sampler to draw samples from this posterior.
\begin{small}
\begin{verbatim}
fns = prior_fn, simulator_fn
model = NRE(fns, make_model())

data, _ = model.simulate_data(jr.PRNGKey(1), n_simulations=10_000)
params, _ = model.fit(jr.PRNGKey(2), data=data)

inference_results, _ = model.sample_posterior(jr.PRNGKey(3), params, y_observed)
\end{verbatim}
\end{small}

\subsection{Neural posterior estimation}
To construct an algorithm based on neural posterior estimation, a user has to construct a conditional density estimator, like a normalizing flow. Below, we provide an example how a normalizing flow that models the approximate posterior can be constructed using the {Python} package \texttt{surjectors}.

\begin{small}
\begin{verbatim}
from surjectors import (
    Chain,
    MaskedAutoregressive,
    Permutation,
    ScalarAffine
    TransformedDistribution,
)
from surjectors.nn import MADE
from surjectors.util import unstack

def make_flow(dim):
    def _bijector_fn(params):
        means, log_scales = unstack(params, -1)
        return surjectors.ScalarAffine(means, jnp.exp(log_scales))

    def _flow(method, **kwargs):
        layers = []
        order = jnp.arange(dim)
        for i in range(5):
            layer = MaskedAutoregressive(
                bijector_fn=_bijector_fn,
                conditioner=MADE(
                    dim, [64, 64], 2,
                    w_init=hk.initializers.TruncatedNormal(0.001),
                    b_init=jnp.zeros,                    
                ),
            )
            order = order[::-1]
            layers.append(layer)
            layers.append(Permutation(order, 1))
        chain = Chain(layers)

        base_distribution = tfd.Independent(
            tfd.Normal(jnp.zeros(dim), jnp.ones(dim)),
            1,
        )
        td = TransformedDistribution(base_distribution, chain)
        return td(method, **kwargs)

    td = hk.transform(_flow)
    return td
\end{verbatim}
\end{small}

With the normalizing flow, inference is straight-forward using the same steps as with any other neural SBI algorithm.
\begin{small}
\begin{verbatim}
fns = prior_fn, simulator_fn
model = NPE(fns, make_flow(2))

data, _ = model.simulate_data(jr.PRNGKey(1), n_simulations=10000)
params, _ = model.fit(jr.PRNGKey(2), data=data)

inference_results, _ = model.sample_posterior(jr.PRNGKey(3), params, y_observed)
\end{verbatim}
\end{small}

\subsection{Neural likelihood estimation}
Here, we infer the posterior distribution using a mixture density network (MDN) as a conditional density estimator. \texttt{sbijax} offers functionality to construct a MDN automatically using \texttt{make\_mdn}, but as illustration we show the approach below (even though it requires in-depth knowledge of \texttt{sbijax}, \texttt{Haiku} and density estimation).
\begin{small}
\begin{verbatim}
def make_mdn():
    @hk.without_apply_rng
    @hk.transform
    def mdn(method, y, x):
        n = x.shape[0]
        hidden = hk.nets.MLP(hidden_sizes, activation=activation, activate_final=True)(x)
        logits = hk.Linear(n_components)(hidden)
        mu_sigma = hk.Linear(n_components * n_dimension * 2)(hidden)
        mu, sigma = jnp.split(mu_sigma, 2, axis=-1)

        mixture = tfd.MixtureSameFamily(
            tfd.Categorical(logits=logits),
            tfd.MultivariateNormalDiag(
                mu.reshape(n, n_components, n_dimension),
                jnp.exp(sigma.reshape(n, n_components, n_dimension)),
            )
        )
        return mixture.log_prob(y)
    return mdn
\end{verbatim}
\end{small}

The approximate posterior is then inferred as before.
\begin{small}
\begin{verbatim}
fns = prior_fn, simulator_fn
model = FMPE(fns, make_model(2))

data, _ = model.simulate_data(jr.PRNGKey(1), n_simulations=10000)
params, _ = model.fit(jr.PRNGKey(2), data=data)

inference_results, _ = model.sample_posterior(jr.PRNGKey(3), params, y_observed)
\end{verbatim}
\end{small}

\section{Additional material for the EEG use case}
\label{app:real-data}

\begin{figure}[h!t]
    \centering
    \includegraphics[width=1\textwidth]{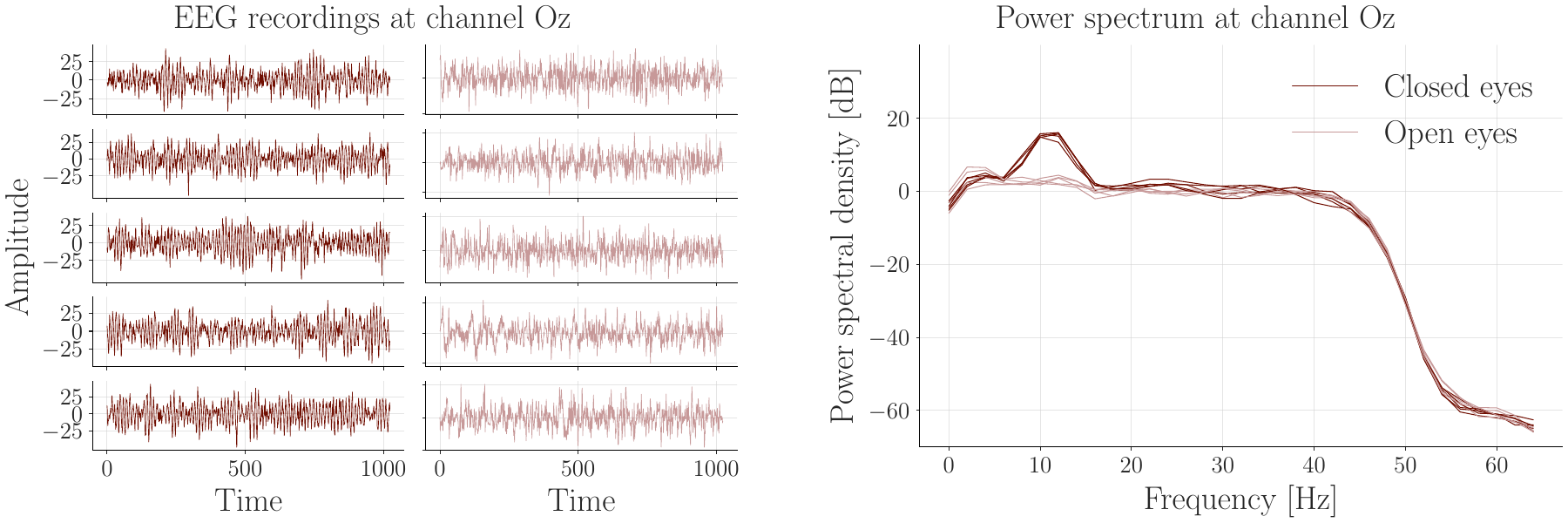}
    \caption{All EEG trajectories for the analyzed test subject. While in the time-domain the measurements for closed- and opened-eyes states show some diversity (left figure), in the frequency-domain the trajectories for a specific eyes state are very similar (right figure). Consequently, choosing a specific pair of observations is uncritical for this use case.}
    \label{fig:eeg-raw-full}
\end{figure}

We demonstrate \texttt{sbijax} using EEG recordings from \citet{cattan2018}. We focus on measurements from the Oz channel which is placed near the visual cortex and, consequently, relevant for the analysis of the open- and closed-eyes states EEGs. Following \citet{rodrigues2021hnpe}, the signals are filtered between $3Hz$ and $40Hz$ and then resampled at $128Hz$. The extracted signals are $8$ seconds long and consist of $1025$ time points per recording. Figure~\ref{fig:eeg-raw-full} shows the entire set of recordings for a specific test subject. The power spectrum on the right is represented in decibels and computed via
\begin{align}
    10 \text{log}_{10} \left( \text{psd}(y) \right)
\end{align}
where $\text{psd}$ is the power spectrum computed via Welch's algorithm using \texttt{scipy} \citep{virtanen2020scipy}.

As a simulator function, which is required to train neural SBI or ABC methods, we use the stochastic Jansen-Rit neural mass model. The JRNMM has, in comparison to the non-stochastic version, additional parameters $\mu, \sigma_3, \sigma_4, \sigma_5$ which collectively govern the amplitude and oscillations of the SDE. It has the following mathematical expression:
\begin{equation*}
\begin{split}
\mathrm{d}Y_0(t) \ &= \ Y_3(t)\mathrm{d}t \\
\mathrm{d}Y_1(t) \ &= \ Y_4(t)\mathrm{d}t \\
\mathrm{d}Y_2(t) \ &= \ Y_5(t)\mathrm{d}t \\
\mathrm{d}Y_3(t) \ &= \ \bigg[ Aa \Big[  \text{sigm}\Big(Y_1(t) - Y_2(t) \Big)  \Big] - 2aY_3(t) - a^2Y_0(t) \bigg]  \mathrm{d}t +\sigma_3 \mathrm{d}W_3(t)\\
\mathrm{d}Y_4(t) \ &= \ \bigg[ Aa \Big[ \mu + C_2\text{sigm}\Big(C_1 Y_0(t) \Big) \Big] - 2aY_4(t) - a^2Y_1(t) \bigg]  \mathrm{d}t +\sigma_4 \mathrm{d}W_4(t)\\
\mathrm{d}Y_5(t) \ &= \ \bigg[ Bb \Big[ C_4\text{sigm}\Big(C_3 Y_0(t) \Big) \Big] - 2bY_5(t) - b^2Y_2(t)  \bigg]  \mathrm{d}t +\sigma_5 \mathrm{d}W_5(t)\\
\end{split}
\end{equation*}
where
\begin{equation*}
    \text{sigm}(y) = \frac{v_{\text{max}}}{1 + \exp(r(v_0 - y))}
\end{equation*}
The hyperparameters are chosen as in previous work where their interpretations can also be found \citep{ableidinger2017stochastic}. Concretely, we use $A=3.25$, $B=22$, $a=100$, $b=50$, $v_{\text{max}}=5$, $v_0=6$, $r=0.56$, $\sigma_3=0.01$ and $\sigma_5=1$.

To infer the posterior distribution for a specific observation, we first simulate $100\ 000$ trajectories from the JRNMM model. Instead of solving the SDE using a conventional solver, like an Euler solver, we refer to the same Strang-splitting method as in \citet{buckwar2020spectral}. The {Python} implementation of the simulator and prior functions can be found in the supplementary code. Calling the simulator with a sample from the prior yields a data set of pairs $\{(y_n, \theta_n)\}_{n=1}^{N=100\ 000}$. We then compute summaries $s_n$ for each trajectory via the power spectral density using the implementation of the Welch algorithm from \texttt{scipy}. Following \citet{rodrigues2021hnpe}, we compute 33 summary statistics, as shown in the {Python} code below:
\begin{small}
\begin{verbatim}
def summarize(y, n_summaries=33, fs=128):
    _, summaries = welch(y, fs=fs, nperseg=2 * (n_summaries - 1), axis=1)
    return summaries

#  see supplementary code for the implementation
(y_train, theta_train), y_closed, y_open = simulate()
summaries_train = summarize(y_train)
\end{verbatim}
\end{small}

Here, we train an amortized posterior estimator, in this case \texttt{NPE}, to approximate the conditional density $\pi(\theta|s)$ where $s$ is a generic summary of an EEG signal. Note that in the \texttt{NPE} constructor below, we provide as a simulator function the null value, \texttt{None}, instead of the simulator function itself.  Since the computation of the synthetic trajectories is fairly computational demanding, we precomputed the $100\ 000$ trajectories and use them instead of simulating them during training. A scenario like this, where precomputed data is available and running the simulator itself is extremely costly, is found frequently in applied sciences, such as astrophysics, and we want to highlight that \texttt{sbijax} naturally supports this scenario.

\begin{small}
\begin{verbatim}
import optax

from jax import random as jr
from sbijax import NPE
from sbijax.nn import make_maf

n_dim_data = 33 # number of summary statistics from prior literature
n_dim_theta = 4 # posterior dimensionality
n_layers, hidden_sizes = 10, (64, 64)
neural_network = make_maf(n_dim_theta, n_layers, hidden_sizes=hidden_sizes)

fns = prior_fn, None
estim = NPE(fns, neural_network)

data = {"y": summaries_train, "theta": theta_train}
params, info = estim.fit(
    jr.PRNGKey(1),
    data=data,
    optimizer=optax.adam(0.0001),
    n_early_stopping_delta=0.01,
    n_early_stopping_patience=10,
)
\end{verbatim}
\end{small}

Having trained the estimator, we can infer the posterior distribution of the measurements \texttt{y\_closed} and \texttt{y\_open} using:
\begin{small}
\begin{verbatim}
posteriors = []
for y in [y_closed, y_open]:
    posterior, _ = estim.sample_posterior(
        jr.PRNGKey(2), 
        params,
        observable=summarize(y),
        n_samples=10_000,
    )
    posteriors.append(posterior)
\end{verbatim}
\end{small}



\end{document}